\documentclass{article}

 \usepackage[eandd, final]{neurips_2026}

\usepackage[utf8]{inputenc} % allow utf-8 input
\usepackage[T1]{fontenc}    % use 8-bit T1 fonts
\usepackage{hyperref}       % hyperlinks
\usepackage{url}            % simple URL typesetting
\usepackage{booktabs}       % professional-quality tables
\usepackage{amsfonts}       % blackboard math symbols
\usepackage{nicefrac}       % compact symbols for 1/2, etc.
\usepackage{microtype}      % microtypography
\usepackage{xcolor}         % colors

\usepackage{amsmath}
\usepackage{multirow}
\usepackage{lipsum}
\usepackage{graphicx}
\usepackage{wrapfig}
\usepackage{tabularx}
\usepackage{tcolorbox}
\usepackage{colortbl}
\usepackage{makecell}
\usepackage{subcaption}
\usepackage{wrapfig}
\usepackage{bm}

\definecolor{c1}{RGB}{192, 0, 0}
\definecolor{c2}{RGB}{0, 112, 192}
\definecolor{c3}{RGB}{0, 0, 0}

\newcommand{\first}[1]{\textbf{{\color{c1}#1}}}
\newcommand{\second}[1]{{\textbf{\color{c2}#1}}}
\newcommand{\third}[1]{{\textbf{\color{c3}#1}}}

\newtcolorbox{observationbox}{
    colback=white,
    colframe=black,
    boxrule=0.8pt,
    arc=3pt,
    left=8pt,
    right=8pt,
    top=5pt,
    bottom=5pt,
    before skip=6pt,
    after skip=6pt,
}

\newcounter{counter}
\newenvironment{observation}{
    \refstepcounter{counter}
    \begin{observationbox}
    \textbf{Obs. \thecounter.}\ 
}{
    \end{observationbox}
}

\newcommand{\fullrowgray}{%
    \noalign{%
        \begingroup
        \color{gray!20}%
        \hrule height 0pt
        \kern-\dimexpr\baselineskip\relax
        \hrule height \baselineskip
        \endgroup
    }%
}

\title{
SGNNBench: A Holistic Evaluation of Spiking Graph Neural Networks on Large-scale Graphs
}

\author{%
  Huizhe Zhang$^{1}$, 
  Jintang Li$^{2}$, 
  Yuchang Zhu$^{1}$, 
  Huazhen Zhong$^{1}$, 
  Liang Chen$^{1,}$\thanks{Corresponding author.} \\
  $^{1}$Sun Yat-sen University,
  $^{2}$Xiamen University \\
  \texttt{\{zhanghzh33, zhuych27, zhonghzh9\}@mail2.sysu.edu.cn} \\
  \texttt{edisonlee@xmu.edu.cn},
  \texttt{chenliang6@mail.sysu.edu.cn}\\
}

\begin{document}

\maketitle

\begin{abstract}
Graph Neural Networks (GNNs), as representative deep models designed for graph data, effectively learn graph topological information and push the performance boundaries across various graph tasks. However, the computational and memory burden of GNNs poses significant challenges for scaling to large real-world graphs.
Spiking Graph Neural Networks (SGNNs), which integrate biologically plausible learning via unique spike-based neurons, have emerged as a promising energy-efficient alternative. Different layers communicate with sparse and binary spikes, which facilitates computation and storage of intermediate graph representations. Despite the proliferation of SGNNs proposed in recent years, there is no systematic benchmark to explore the basic design principles of these brain-inspired networks on graph data.
To bridge this gap, we present \textbf{SGNNBench} to quantify progress in the field of SGNNs. Specifically, we thoroughly investigate the design space of SGNNs to facilitate the development of a general SGNN paradigm. Regarding efficiency, we empirically compare these baselines w.r.t. model size, memory usage and theoretical energy consumption to reveal previously overlooked energy bottlenecks. Finally, SGNNBench establishes a unified and reliable benchmark to comprehensively evaluate 10 state-of-the-art SGNNs across 24 datasets covering diverse graph settings and tasks. Source code is available at \url{https://github.com/Zhhuizhe/SGNNBench}.
\end{abstract}

\section{Introduction}\label{sec:1}
Graph Neural Networks (GNNs) have become dominant deep representation learning methods designed for graph data (e.g., social networks, biological systems, recommendation systems) \cite{kipf2016semi,vaswani2017attention,wu2020comprehensive}. As the core feature of GNNs, the message passing mechanism captures the graph topology and node features to learn more expressive node representations. Numerous empirical experiments show that GNNs have brought promising performance gains and become the standard graph mining approach across various graph tasks \cite{zhu2024one, zhu2024fair, li2024state}.
However, the nature of preserving the integral graph structures incurs considerable computational and storage overhead on large-scale graphs. For example, training a GNN-based method on a graph with billions of nodes and edges will require up to three days and 384 GB of memory on a 16-GPU cluster \cite{ying2018graph}. The high energy consumption caused by the sheer number of nodes and edges has exposed an efficiency bottleneck, which limits the deployment of GNN-based methods for real-world graphs.

\begin{figure}[t]
    \centering
    \includegraphics[width=\linewidth]{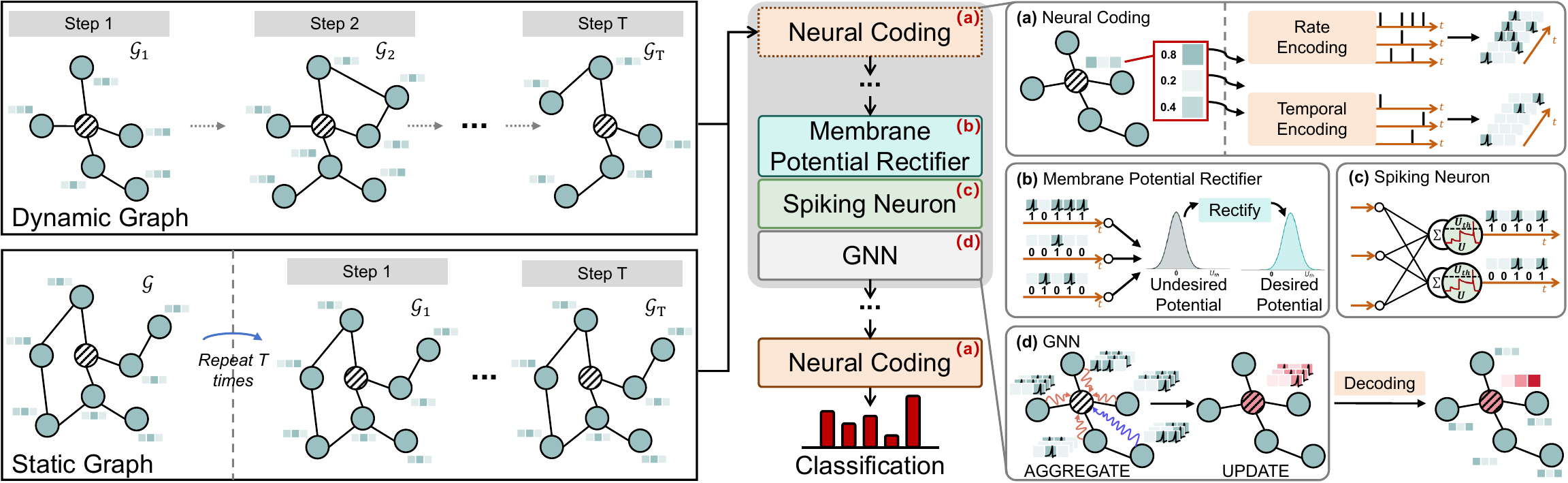}
    \caption{The design space of existing SGNNs. Common architectural components of SGNNs can be categorized into four key modules: \textbf{(a)} neural coding scheme, \textbf{(b)} membrane potential rectifier, \textbf{(c)} spiking neuron, and \textbf{(d)} GNN layer.} \label{fig:1}
\end{figure}

With the rise of neuromorphic chips, Spiking Neural Networks (SNNs) have emerged as energy-efficient alternatives to traditional Artificial Neural Networks (ANNs), converting real-valued signals into sparse, single-bit spike trains via biologically inspired neurons \cite{eshraghian2023training}. It endows the neural networks with advantages in storage efficiency and computational simplicity. We summarize the architectural design space of existing SGNNs in Figure~\ref{fig:1}.
Although recent studies have demonstrated the potential of SGNNs for energy-saving graph representation learning, the lack of a standardized evaluation pipeline hampers efforts to develop a general spiking-driven GNN paradigm. Existing comparisons based on outdated baselines obscure the unresolved challenges of SGNNs, while models tailored to specific graph properties hinder the development of generalizable SGNNs. It makes existing studies more of an intellectual endeavour than guidance for practical improvements.
Here, we summarize the pressing research questions for SGNNs:
\textbf{(RQ1)} How much progress has been made by SGNNs compared with classic full-precision GNNs?
\textbf{(RQ2)} How efficient are these SGNNs in terms of energy consumption, and which design achieves the sweet spot between effectiveness and efficiency? 
\textbf{(RQ3)} Are commonly used components designed for full-precision GNNs equally effective for SGNNs? 

To answer these questions, we revisit existing state-of-the-art SGNNs and provide a comprehensive benchmark study referred to as \textbf{SGNNBench}. SGNNBench aims to provide a fair and consistent comparison among different SGNN baselines and examine the performance gaps between SGNNs and classic full-precision GNNs. For SGNN baselines, we conducted a thorough energy efficiency analysis using metrics such as memory usage, model size and theoretical energy consumption. In the ablation study, we systematically analyze the pros and cons of existing spiking neuron variants and rectification operations. Overall, the results reveal the overlooked high-energy operations in current SGNNs and provide deeper insights for future model design. The contributions of this work can be summarized as follows:

\begin{itemize}
\item As far as we know, SGNNBench is the first benchmark study that conducts a comprehensive comparison of 10 SGNN baselines with 16 classic full-precision GNNs on 24 datasets. 
\item We conduct an energy efficiency analysis to evaluate the time and space complexity of SGNN branches via memory footprint, model size and theoretical energy consumption.
\item We conduct an ablation study on advanced spiking neurons, normalization techniques and neural network components to yield valuable insights. The results indicate that the effectiveness of conventional GNN architectural components has been largely overlooked in the design of SGNNs. With proper integration of these components, SGNNs can further narrow the performance gap with their full-precision counterparts.
\end{itemize}

\section{Design Space For Spiking Graph Neural Networks}\label{sec:2}

\subsection{Graph Neural Network}
Given a graph with a set of nodes $\mathcal{V}$ and edges $\mathcal{E}$, it can be represented as $\mathcal{G}=(\mathcal{V},\mathcal{E})$. The node features are formally defined as $\bm{X} \in \mathbb{R}^{N \times d}$, where $N$ is the number of nodes and $d$ is the dimensionality of features. Besides, $\bm{A} \in \{0,1\}^{N \times N}$ is the adjacency matrix, where each element indicates whether a pair of nodes is connected. 
The mainstream GNNs are built upon a key mechanism called the message passing mechanism. For a node $u\in \mathcal{V}$, its hidden embedding $\bm{h}_u\in\mathbb{R}^{d^{\prime}}$ is updated by aggregating the information from its immediate neighborhood $\mathcal{N}(u)$. Each node embedding progressively incorporates more information from further reaches of the graph via multiple iterations. The whole process can be expressed as follows:
\begin{align}
    \bm{m}^l_{\mathcal{N}_{(u)}}&=\operatorname{AGGREGATE}(\{\bm{h}^l_v, \; \forall v\in\mathcal{N}(u)\}), \\
    \bm{h}^{l+1}_u&=\operatorname{UPDATE}(\bm{h}^{l}_u, \; \bm{m}^l_{\bm{N}_{(u)}}),
\end{align}
where $\bm{h}^l_u$ denotes the embedding of node $u$ in the $l$-th layer. The $\operatorname{AGGREGATE}$ function is responsible for gathering and combining information from a node's neighbors. The $\operatorname{UPDATE}$ function is used to modify the node's feature vector. To address the issue concerning under-reaching, a line of research has explored higher-order interaction operators to capture long-range dependencies \cite{morris2019weisfeiler, wang2023mathscrnwl, li2023local}. Furthermore, Graph Transformers (GTs) aggregate information over all node pairs through self-attention. For simplicity, we omit the scaling factor and the multi-head mechanism. The self-attention block can be reformulated under the AGGREGATE-UPDATE paradigm as follows:
\begin{align}
    \bm{q}_u&=\bm{W}_Q\bm{h}_u,\;\bm{k}_u=\bm{W}_K\bm{h}_u,\;\bm{v}_u=\bm{W}_V\bm{h}_u \\
    \bm{m}_j&=\operatorname{ATTN}({\bm{h}_u}, \bm{h}_j)=\frac{\operatorname{exp}(\bm{q}_u \bm{k}^T_j)}{\sum^N_{w=1}{\operatorname{exp}(\bm{q}_u \bm{k}^T_w)}} \bm{v}_j, \\
    \bm{h}^{l+1}_u&=\operatorname{MLP}(\operatorname{LayerNorm}(\bm{h}^l_u+\sum^N_{j=1}\bm{m}^l_j)).
\end{align}

\subsection{Spiking Neuron}
Unlike Artificial Neural Networks (ANNs), biological neurons in SNNs communicate with action potentials (also referred to as spikes). The Leaky Integrate-and-Fire (LIF) neurons \cite{salinas2002integrate}, as simplified phenomenological models, provide an abstract description of action potentials in neurons. The dynamics of the membrane potential in LIF neurons can be modeled as an RC circuit:
\begin{align}\label{eq:1}
\tau\frac{\mathrm{d}U}{\mathrm{d}t}=-(U^t - U_{reset})+RI^t,
\end{align}
where $\tau$ denotes the time constant of the circuit. By solving the linear ODE via the forward Euler method, we can abstract the process of spike generation and propagation into three steps: (i) \textbf{Integrate:} the neuron integrates synaptic inputs from other neurons or external current $\bm{I}\in\mathbb{R}^{N \times d^{\prime}}$ to charge its cell membrane. (ii) \textbf{Fire:} when the membrane potential reaches a pre-defined threshold $\bm{U}_{th}\in\mathbb{R}^{N \times d^{\prime}}$, the neuron fires spikes $\bm{S} \in \{0, 1\}^{N \times d^{\prime}}$. (iii) \textbf{Reset:} the membrane potential of the neuron will be reset to $\bm{U}_{reset}\in\mathbb{R}^{N \times d^{\prime}}$ after firing. The neuronal dynamics are formulated as follows:
\begin{align}\label{eq:2}
    \textbf{Integrate:} & \; \bm{U}^t=\Psi(\bm{U}^{t-1}, \bm{I}^t)=\bm{U}^{t-1} + \frac{1}{\tau}(\bm{I}^t-(\bm{U}^{t-1}-\bm{U}_{reset})), \\
    \textbf{Fire:} & \; 
    \bm{S}^t = \Theta(\bm{U}^t-\bm{U}_{th})=\left\{
    \begin{array}{ll}
         1,& \bm{U}^t - \bm{U}_{th} \geq 0 \\
         0,& \text{otherwise}
    \end{array},\right.  \\
    \textbf{Reset:} & \; \bm{U}^{t} = \bm{U}^t \odot (\bm{J} - \bm{S}^t) + \bm{U}_{reset} \odot \bm{S}^t,   
\end{align}
where $\bm{J}$ is an $N \times d^{\prime}$ all-ones matrix and $\odot$ denotes the Hadamard product. $\bm{U}^t$ and $\bm{I}^t$ are the membrane potential and the pre-synaptic input at time step $t$. $\tau$ is a decay factor of the membrane potential. $\Theta(\cdot)$ is the Heaviside step function and $\Psi(\cdot)$ is the membrane potential update function. 
Although the spiking neuron variants have been extensively explored in other domains \cite{yao2022glif, guo2024ternary, huang2024clif}, there remains considerable scope for investigating their applicability and developing graph-specific neurons to learn more expressive spiking graph representations \cite{chen2025signn, jahin2026chronospike}.

\subsection{Neural Coding Scheme} 
Neural coding schemes aim to convert input features into spike signals for transmission to excitatory neurons. There are two common coding schemes: rate coding and temporal coding. 
Rate coding transforms the intensity or magnitude of an input signal into the frequency of spikes. Higher input values may lead to higher firing rates or spike counts.
For temporal coding, the input intensity will be converted to the spike time. The larger input current leads the membrane potential to reach $U_{th}$ more quickly, leading to faster spike generation: 
\begin{align}
    \textbf{Rate\ coding:} & \; P(\bm{s}=1)=\bm{h}=1-P(\bm{s}=0), \\
    \textbf{Temporal\ coding:} & \;  t(\bm{h})=\left\{
    \begin{array}{ll}
    \tau \ln(\frac{\bm{h}}{\bm{h}-\bm{u}_{th}}),& \ \bm{h}>\bm{u}_{th} \\
    \infty,& \ \text{otherwise}
    \end{array}
    \right.,
\end{align}
Since some SGNNs built upon temporal coding \cite{dold2021spike, dold2022relational, xiao2024temporal} are designed exclusively for inductive knowledge graph completion, they are excluded from the following comparison.

\subsection{Membrane Potential Rectifier} 
For spiking neurons, the dynamics of the membrane potential determine when and how neurons fire. Inappropriate membrane potential distribution may result in excessive or insufficient spiking. To address this, the membrane potential rectifier is introduced to regulate potential values and shape spike activity in a controlled, task-adaptive manner \cite{zheng2021going, guo2022recdis}. We define the \textit{quantization distortion} between a full-precision representation vector and the corresponding spiking vector as follows:
\begin{align}
D=\frac{1}{N}\sum_{u\in\mathcal{V}}\sum_{t}\|\bm{h}^t_u - \bm{s}^t_u\|_2=\frac{1}{N}\sum_{u\in\mathcal{V}}\sum_{t}\|\bm{h}^t_u - \Theta(\phi(\bm{h}^t_u; \theta))\|_2,
\end{align}
where $\phi_{\theta}(\cdot)$ is the rectifying function used to regulate the membrane potential. 
Membrane potential rectifiers include initialization and rectification to mitigate undesired voltage shifts and reduce quantization loss \cite{shen2024rethinking}. For SGNNs, these dynamics are also linked to graph properties. For example, spatial–temporal normalization regulates the membrane potential of nodes across spatial and temporal dimensions \cite{xu2021exploiting}, while degree-aware threshold adaptation can be viewed as post-rectification that accounts for node degree to produce more expressive spiking patterns \cite{wang2024degree}.

\begin{figure*}[t]
    \centering
    \includegraphics[width=\linewidth]{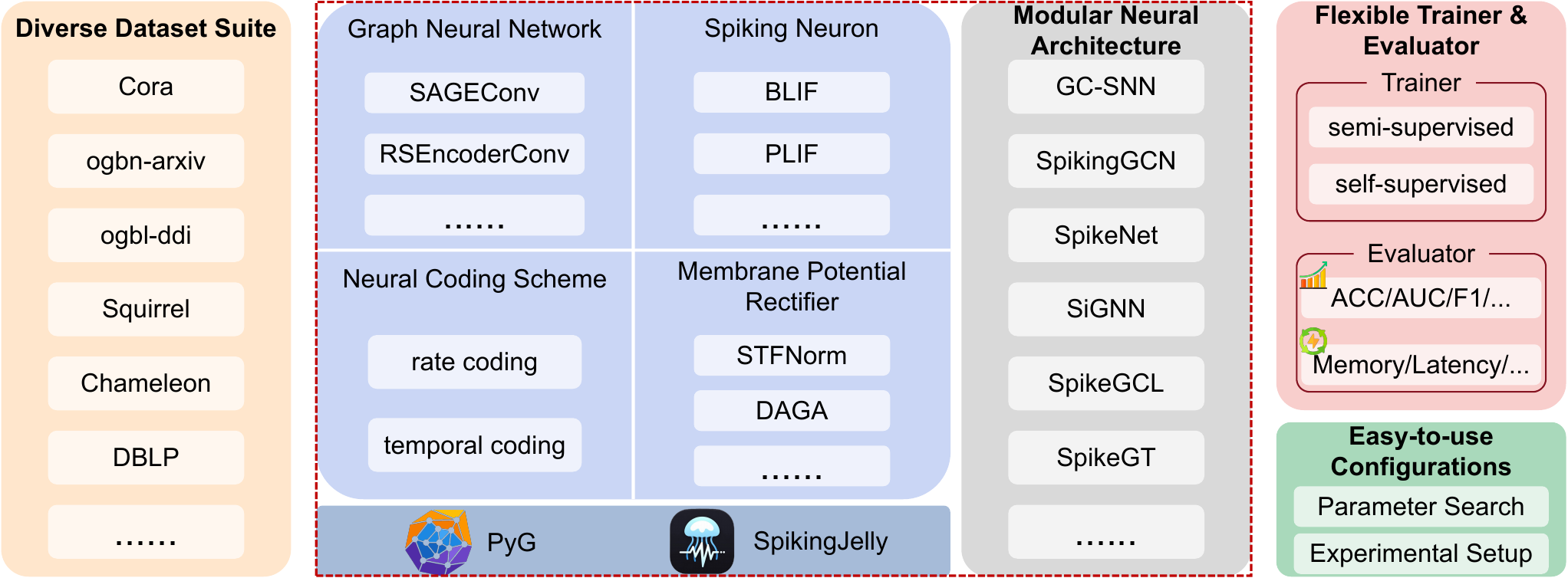}
    \caption{The overview of SGNNBench.} \label{fig:overview}
    \vspace{-10px}
\end{figure*}

\section{SGNNBench}
Although SGNNs benefit from the computational and memory efficiency of spikes, existing evaluations are often conducted on non-standardized pipelines, outdated GNN baselines and inconsistent implementations. This has hampered the real progress of SGNNs. Moreover, many existing SGNNs are tailored to narrow domains, which limits the development of a general paradigm. To date, there has been no benchmarking effort that systematically reveals the efficiency gains brought by spikes while providing a holistic understanding of the SGNN design space.

SGNNBench is proposed to address the above critical gap by establishing the first holistic benchmark of SGNNs. As depicted in Figure~\ref{fig:overview}, SGNNBench is built upon four key characteristics: 
\textbf{(i) Diverse Dataset Suite}, covering graph datasets with diverse scales, properties and tasks; 
\textbf{(ii) Modular Neural Architecture}, organizing reusable network components according to the design space of SGNNs; 
\textbf{(iii) Flexible Trainer and Evaluator}, supporting graph-specific learning paradigms; 
\textbf{(iv) Easy-to-use Configurations}, providing standardized and easy-to-use experimental configurations. 
Overall, SGNNBench aims to establish unified evaluation standards and foster the development of more generalizable and practical SGNNs.

\paragraph{Diverse Dataset Suite.} 
SGNNBench includes 24 datasets spanning diverse graph settings and tasks (e.g., static/dynamic graphs, homogeneous/heterogeneous graphs, node classification and link prediction) to comprehensively evaluate the generalization capability of SGNN designs.

\paragraph{Modular Neural Architecture.}
SGNNBench systematically abstracts the design space of SGNNs and organizes reusable network components in a modular framework. Based on this unified design space, the benchmark integrates 10 state-of-the-art SGNNs together with stronger full-precision GNN baselines \cite{luo2024classic} for fair and standardized comparisons.

\paragraph{Flexible Trainer and Evaluator.} 
SGNNBench provides a flexible training and evaluation engine tailored for graph learning, supporting semi-supervised paradigms in both transductive and inductive settings. SGNNBench further reports both task-specific performance metrics and efficiency-related metrics for comprehensive evaluation.

\paragraph{Easy-to-use Configurations.} 
SGNNBench provides standardized configuration files for experiment management and result reproduction, enabling transparent, easy-to-use and reproducible evaluations across different SGNN settings.

\section{Benchmarking Over Effectiveness (RQ1)}\label{sec:exp_effectiveness}
\subsection{Experimental Setup for Comparison}
In this section, we conduct a series of experiments to comprehensively measure the performance gaps between state-of-the-art SGNNs and classic full-precision GNNs. SGNNBench implements 10 SGNN baselines to conduct comparison experiments on 24 datasets following the previous task setting \cite{li2023scaling, luo2024classic, gastinger2024tgb}. For those methods lacking the official code, we provide our own implementation based on the original papers. 
The hyperparameter search is performed to obtain the optimal hyperparameter combinations for each baseline. We run each baseline five times on every dataset with different random seeds, and the mean and standard deviation are reported as evaluation metrics. More details about experimental settings are provided in Appendix~\ref{sup:exp_setting}.

\subsection{Comparison on Static Graphs}\label{sec:3_2}

\paragraph{Node classification.} The experimental results on homophilic and heterophilic graphs are demonstrated in Table~\ref{tab:homo_nc} and Table~\ref{tab:hetero_nc}, respectively, with additional results provided in Appendix~\ref{sup:extra_exp_static}.
As shown in Table~\ref{tab:homo_nc}, we adopt the average classification accuracy as the evaluation metric. SGNNs have achieved approximately $92\%\sim99\%$ of the performance of full-precision GNNs on these datasets. Across 9 datasets, the average performance gap between the best-performing full-precision GNNs and spiking counterparts is approximately 2.69 percentage points. Due to the binarized spiking node representations, which can be seen as the extreme case of quantization, the performance degradation of SGNNs is expected and acceptable. We believe that SGNNs are strong contenders for energy-efficient graph representation learning, offering a promising balance between accuracy and efficiency.

\textbf{We take the lead in comprehensively evaluating the performance of SGNNs on heterophilic graphs.} We find that most existing SGNNs are tailored for homophilic settings. SGNNs struggle to match their full-precision counterparts on heterophilic graphs with diverse structural properties and natures, especially on Roman-Empire. This graph with a chain-like structure has the largest diameter across all datasets. We notice that deeper full-precision GNNs (typically with $8\sim10$ layers) tend to achieve the best performance. For SGNNs, increasing network depth may exacerbate the issues concerning over-smoothing and quantization loss simultaneously. It hinders the performance gains typically expected from deeper architectures on such graphs and results in a significant performance gap compared to full-precision GNNs on heterophilic graphs.

\begin{observation}
    SGNNs exhibit only marginal performance gaps compared to full-precision GNNs, positioning them as strong contenders for energy-efficient graph representation learning. \textbf{However, the incompatibility between spiking node embeddings and deeper network architectures remains unresolved.} Extending existing SGNNs to heterophilic graphs often leads to more pronounced performance degradation.
\end{observation}

\begin{table}[t]
    \scriptsize
    \tabcolsep=1.8px
    \centering
    \caption{Node classification (\%) on homophilic graphs. $^*$ denotes our implementation. The \first{best}, \second{second-best} and \third{third-best} results among SGNNs are highlighted.}\label{tab:homo_nc}
    \begin{tabular}{lccccccccccc}
        \toprule
        \textbf{Models} & \textbf{Cora} & \textbf{CiteSeer} & \textbf{PubMed} & \textbf{Computers} & \textbf{Photo} & \textbf{CS} & \textbf{Physics} & \textbf{ogbn-proteins} & \textbf{ogbn-arxiv} \\
        \midrule
        GCN  & 85.10$_{\pm0.67}$ & 73.14$_{\pm0.67}$ & 81.12$_{\pm0.52}$ & 93.99$_{\pm0.12}$ & 96.10$_{\pm0.46}$ & 96.17$_{\pm0.06}$ & 97.46$_{\pm0.10}$ & 77.29$_{\pm0.35}$ & 73.53$_{\pm0.36}$ \\
        SAGE & 83.88$_{\pm0.65}$ & 72.26$_{\pm0.55}$ & 79.72$_{\pm0.50}$ & 93.25$_{\pm0.14}$ & 96.78$_{\pm0.23}$ & 96.38$_{\pm0.11}$ & 97.19$_{\pm0.05}$ & 82.21$_{\pm0.20}$ & 73.00$_{\pm0.27}$ \\
        GAT  & 84.46$_{\pm0.55}$ & 72.22$_{\pm0.84}$ & 80.28$_{\pm0.64}$ & 94.09$_{\pm0.37}$ & 96.60$_{\pm0.33}$ & 96.21$_{\pm0.14}$ & 97.25$_{\pm0.06}$ & 85.01$_{\pm0.44}$ & 73.30$_{\pm0.34}$ \\
        SGC  & 79.96$_{\pm1.67}$ & 67.06$_{\pm0.59}$ & 76.94$_{\pm0.67}$ & 91.36$_{\pm0.41}$ & 93.36$_{\pm0.26}$ & 95.16$_{\pm0.18}$ & 96.23$_{\pm0.18}$ & 77.00$_{\pm1.68}$ & 72.05$_{\pm0.60}$ \\
        \rowcolor{gray!20}
        Average & 83.35 & 71.17 & 79.52 & 93.17 & 95.71 & 95.98 & 97.03 & 80.37 & 72.97 \\
        \midrule
        GPRGNN     & 83.00$_{\pm1.44}$ & 71.86$_{\pm1.17}$ & 78.78$_{\pm3.32}$ & 92.51$_{\pm0.28}$ & 95.65$_{\pm0.36}$ & 95.92$_{\pm0.20}$ & 97.22$_{\pm0.07}$ & 75.68$_{\pm0.49}$ & 71.10$_{\pm0.12}$ \\
        ChebNet\MakeUppercase{\romannumeral 2}  & 79.88$_{\pm1.52}$ & 69.66$_{\pm0.98}$ & 78.48$_{\pm3.18}$ & 92.79$_{\pm0.28}$ & 95.62$_{\pm0.48}$ & 95.82$_{\pm0.29}$ & 97.15$_{\pm0.12}$ & 72.41$_{\pm0.27}$ & 72.32$_{\pm0.23}$ \\
        SGFormer   & 84.82$_{\pm0.85}$ & 72.72$_{\pm1.15}$ & 80.60$_{\pm0.49}$ & 92.42$_{\pm0.66}$ & 95.58$_{\pm0.36}$ & 95.71$_{\pm0.24}$ & 96.75$_{\pm0.26}$ & 82.13$_{\pm0.34}$ & 73.14$_{\pm0.16}$ \\
        Polynormer & 83.43$_{\pm0.89}$ & 72.19$_{\pm0.83}$ & 79.35$_{\pm0.73}$ & 93.78$_{\pm0.01}$ & 96.57$_{\pm0.23}$ & 95.42$_{\pm0.19}$ & 97.18$_{\pm0.11}$ & 79.53$_{\pm0.67}$ & 73.40$_{\pm0.22}$ \\
        \rowcolor{gray!20}
        Average & 82.78 & 71.61 & 79.30 & 92.88 & 95.86 & 95.72 & 97.07 & 77.44 & 72.49 \\
        \midrule
        GC-SNN$^*$ & \third{80.00$_{\pm0.78}$} & 67.20$_{\pm0.68}$ & \second{77.35$_{\pm0.65}$} & \second{91.03$_{\pm0.88}$} & 92.79$_{\pm0.68}$ & 93.68$_{\pm1.71}$ & \second{96.67$_{\pm0.39}$} & OOM & 66.83$_{\pm0.30}$ \\
        GA-SNN$^*$ & 76.87$_{\pm4.00}$ & 64.80$_{\pm1.49}$ & 76.10$_{\pm2.16}$ & 88.15$_{\pm0.71}$ & 92.61$_{\pm1.72}$ & 90.31$_{\pm0.41}$ & 95.51$_{\pm0.20}$  & OOM & OOM \\
        SpikingGCN & 74.40$_{\pm1.35}$ & 63.20$_{\pm1.93}$ & 74.03$_{\pm1.18}$ & 85.64$_{\pm1.01}$ & 92.18$_{\pm0.75}$ & 91.71$_{\pm0.24}$ & 95.30$_{\pm0.16}$ & 71.22$_{\pm0.73}$ & 63.86$_{\pm0.72}$ \\
        SpikeNet   & 77.90$_{\pm2.32}$ & 64.82$_{\pm1.33}$ & 75.30$_{\pm1.07}$ & 89.62$_{\pm0.15}$ & \second{94.42$_{\pm0.20}$} & \second{95.05$_{\pm0.35}$} & \first{96.99$_{\pm0.15}$} & \third{73.60$_{\pm0.98}$} & 66.81$_{\pm0.12}$ \\
        SiGNN      & 77.94$_{\pm1.78}$ & 65.93$_{\pm1.16}$ & 75.47$_{\pm1.35}$ & \third{90.61$_{\pm0.07}$} & \first{94.73$_{\pm0.22}$} & \first{95.27$_{\pm0.33}$} & \first{96.99$_{\pm0.10}$} & \second{74.83$_{\pm0.93}$} & \third{68.53$_{\pm0.49}$} \\
        SpikeGCL   & \second{80.43$_{\pm1.10}$} & 66.70$_{\pm0.57}$ & \first{79.40$_{\pm1.02}$} & \first{91.36$_{\pm0.07}$} & 93.75$_{\pm0.34}$ & 93.25$_{\pm0.36}$ & 95.63$_{\pm0.16}$ & OOM & \first{70.85$_{\pm0.31}$} \\
        SpikeGT    & \first{82.43$_{\pm2.10}$} & \first{69.97$_{\pm1.68}$} & \third{76.24$_{\pm2.67}$} & 87.61$_{\pm1.92}$ & 93.70$_{\pm0.17}$ & 93.56$_{\pm0.28}$ & \third{96.51$_{\pm0.13}$} & \first{77.44$_{\pm1.20}$} & \second{70.23$_{\pm0.90}$} \\
        MSG        & 76.30$_{\pm1.27}$ & \second{69.40$_{\pm0.71}$} & 74.35$_{\pm2.54}$ & 87.32$_{\pm0.21}$ & \third{93.86$_{\pm0.27}$} & \third{94.12$_{\pm0.24}$} & \second{96.67$_{\pm0.09}$} & OOM & 67.42$_{\pm0.68}$ \\
        DRSGNN     & 78.98$_{\pm1.46}$ & \third{68.20$_{\pm1.94}$} & 73.78$_{\pm4.40}$ & 87.97$_{\pm0.54}$ & 92.94$_{\pm0.42}$ & 92.63$_{\pm0.77}$ & 95.82$_{\pm0.15}$ & >3days & 66.26$_{\pm0.39}$ \\
        \rowcolor{gray!20}
        Average    & 78.36 & 66.69 & 75.78 & 88.81 & 93.44 & 93.29 & 96.23 & 74.27 & 67.60 \\
        \bottomrule
    \end{tabular}
    \vspace{-10px}
\end{table}

\begin{table}[t]
    \centering
    \scriptsize
    \caption{Node classification (\%) on heterophilic graphs. $^*$ denotes our implementation. The \first{best}, \second{second-best} and \third{third-best} results among SGNNs are highlighted.}\label{tab:hetero_nc}
    \resizebox{\textwidth}{!}{
    \begin{tabular}{lcccccc}
        \toprule
        \textbf{Models}  & \textbf{Squirrel} & \textbf{Chameleon} & \textbf{Amazon-Ratings} & \textbf{Roman-Empire} & \textbf{Minesweeper} & \textbf{Questions} \\
        \midrule
        GCN  & 45.01$_{\pm1.63}$ & 46.29$_{\pm3.40}$ & 53.80$_{\pm0.60}$ & 91.27$_{\pm0.20}$ & 97.86$_{\pm0.24}$ & 79.02$_{\pm0.60}$ \\
        SAGE & 40.78$_{\pm1.47}$ & 44.81$_{\pm4.74}$ & 55.40$_{\pm0.21}$ & 91.06$_{\pm0.27}$ & 97.77$_{\pm0.62}$ & 77.21$_{\pm1.28}$ \\
        GAT  & 41.73$_{\pm2.07}$ & 44.13$_{\pm4.17}$ & 55.54$_{\pm0.51}$ & 90.63$_{\pm0.14}$ & 97.73$_{\pm0.73}$ & 77.95$_{\pm0.51}$ \\
        SGC  & 42.28$_{\pm1.47}$ & 45.57$_{\pm2.00}$ & 51.66$_{\pm0.52}$ & 85.20$_{\pm0.54}$ & 86.13$_{\pm0.88}$ & 77.34$_{\pm0.25}$ \\
        \rowcolor{gray!20}
        Average & 42.45 & 45.20 & 54.10 & 89.54 & 94.87 & 77.88  \\
        \midrule
        H2GCN & 35.10$_{\pm1.15}$ & 26.75$_{\pm3.64}$ & 36.47$_{\pm0.23}$ & 60.11$_{\pm0.52}$ & 89.71$_{\pm0.31}$ & 63.59$_{\pm1.46}$ \\
        GPRGNN & 38.95$_{\pm1.99}$ & 39.93$_{\pm3.30}$ & 44.88$_{\pm0.34}$ & 64.85$_{\pm0.27}$ & 86.24$_{\pm0.61}$ & 55.48$_{\pm0.91}$ \\
        FSGNN & 35.92$_{\pm1.32}$ & 40.61$_{\pm2.97}$ & 52.74$_{\pm0.83}$ & 79.92$_{\pm0.56}$ & 90.08$_{\pm0.70}$ & 78.86$_{\pm0.92}$ \\
        \rowcolor{gray!20}
        Average & 36.66 & 35.76 & 44.7 & 68.29 & 88.68 & 65.98  \\
        \midrule
        GC-SNN$^*$ & \third{43.49$_{\pm1.30}$} & 41.68$_{\pm3.67}$ & 41.82$_{\pm0.28}$ & 55.79$_{\pm0.57}$ & 82.27$_{\pm0.70}$ & 71.36$_{\pm1.10}$ \\
        GA-SNN$^*$ & 40.45$_{\pm0.82}$ & 40.42$_{\pm2.13}$ & 43.83$_{\pm0.64}$ & 65.91$_{\pm1.02}$ & \second{87.92$_{\pm0.27}$} & 65.51$_{\pm0.39}$ \\
        SpikingGCN & 41.49$_{\pm0.87}$ & 40.09$_{\pm1.65}$ & \first{49.47$_{\pm0.49}$} & 65.23$_{\pm1.04}$ & 74.17$_{\pm1.13}$ & \second{76.18$_{\pm0.58}$} \\
        SpikeNet   & 40.00$_{\pm0.45}$ & 42.33$_{\pm3.46}$ & 45.33$_{\pm0.09}$ & \third{71.15$_{\pm0.31}$} & \first{90.42$_{\pm0.17}$} & \third{75.39$_{\pm1.06}$} \\
        SiGNN      & 40.81$_{\pm0.86}$ & 43.20$_{\pm1.78}$ & \second{49.46$_{\pm0.67}$} & \first{73.01$_{\pm0.29}$} & \second{90.38$_{\pm0.31}$} & 74.23$_{\pm0.63}$ \\
        SpikeGCL   & \second{44.38$_{\pm1.56}$} & \third{46.13$_{\pm3.48}$} & \third{46.40$_{\pm0.16}$} & 53.49$_{\pm1.07}$ & 75.31$_{\pm1.27}$ & \first{77.26$_{\pm0.58}$} \\
        SpikeGT    & 38.52$_{\pm0.68}$ & \second{60.83$_{\pm1.80}$} & 46.08$_{\pm0.66}$ & \second{71.86$_{\pm0.55}$} & 87.45$_{\pm0.87}$ & 72.74$_{\pm2.93}$ \\
        MSG        & \first{48.66$_{\pm1.37}$} & \first{63.55$_{\pm1.35}$} & 41.43$_{\pm0.25}$ & 41.96$_{\pm0.70}$ & 82.19$_{\pm0.74}$ & 60.02$_{\pm1.09}$ \\
        DRSGNN     & 41.85$_{\pm2.30}$ & 42.98$_{\pm3.78}$ & 39.77$_{\pm0.15}$ & 26.03$_{\pm2.67}$ & 64.27$_{\pm4.37}$ & 66.25$_{\pm1.55}$ \\
        \rowcolor{gray!20}
        Average    & 42.18 & 46.80 & 44.84 & 58.27 & 81.60 & 70.99 \\
        \bottomrule
    \end{tabular}
    }
    \vspace{-10px}
\end{table}

\paragraph{Link prediction.}
Following the above experimental settings, we report the average AUC, AUPR and Hits@100 over five runs in Table~\ref{tab:homo_lp}. The average performance gap between full-precision GNNs and SGNNs across the five datasets is 3.58 percentage points. As a self-supervised method, SpikeGCL utilizes a contrastive objective to achieve better generalization and consistently outperforms the other SGNN baselines on the reported metrics.
Although the simplified summation aggregator in SpikeNet and SiGNN enables more efficient computation, recursively sampling a fixed number of neighbors bounds the neighborhood size per node. Link prediction requires processing both endpoints of each edge and their sampled neighborhoods. Retaining intermediate embeddings across multiple spiking time steps will incur substantial memory overhead, particularly with large edge batches and multi-hop sampling. These scalability limitations are evident on ogbl-ppa, as most SGNNs run out of memory and the remaining models underperform full-precision GNNs.

\begin{observation}
    For SGNNs, advances in architectural designs and training paradigms also benefit link prediction. However, existing SGNNs still lack scalable neighborhood sampling, efficient storage of intermediate spiking states and advanced spike coding strategies for link prediction in large-scale graphs.
\end{observation}

\begin{table}[tbp]
    \scriptsize
    \tabcolsep=1.8px
    \centering
    \caption{Link prediction (\%) in terms of AUPR, AUC and Hits@100. $^*$ denotes our implementation. The \first{best}, \second{second-best} and \third{third-best} results among SGNNs are highlighted.}\label{tab:homo_lp}
    \begin{tabular}{l|cc|cc|cc|cc|c}
        \toprule
        \multirow{2}{*}{\textbf{Models}} & \multicolumn{2}{c|}{\textbf{Cora}} & \multicolumn{2}{c|}{\textbf{PubMed}} & \multicolumn{2}{c|}{\textbf{ogbl-collab}} & \multicolumn{2}{c|}{\textbf{ogbl-ddi}} & \textbf{ogbl-ppa} \\
        \cmidrule(lr){2-3} \cmidrule(lr){4-5} \cmidrule(lr){6-7} \cmidrule(lr){8-9} \cmidrule(lr){10-10}
        & \textbf{AUPR} & \textbf{AUC} & \textbf{AUPR} & \textbf{AUC} & \textbf{AUPR} & \textbf{AUC} & \textbf{AUPR} & \textbf{AUC} & \textbf{Hits@100} \\
        \midrule
        GCN  & 97.69$_{\pm0.52}$ & 97.89$_{\pm0.34}$ & 98.23$_{\pm0.19}$ & 98.08$_{\pm0.17}$ & 94.35$_{\pm0.18}$ & 96.39$_{\pm0.13}$ & 92.60$_{\pm0.59}$ & 91.39$_{\pm0.59}$ & 16.09$_{\pm0.66}$ \\
        SAGE & 97.37$_{\pm0.57}$ & 97.65$_{\pm0.41}$ & 96.49$_{\pm0.08}$ & 96.86$_{\pm0.09}$ & 92.96$_{\pm0.31}$ & 95.32$_{\pm0.27}$ & 94.65$_{\pm0.17}$ & 93.33$_{\pm0.17}$ & 18.35$_{\pm0.85}$ \\
        GAT  & 97.32$_{\pm0.71}$ & 97.56$_{\pm0.64}$ & 96.50$_{\pm0.19}$ & 97.07$_{\pm0.09}$ & 93.64$_{\pm0.16}$ & 95.97$_{\pm0.10}$ & 95.42$_{\pm0.10}$ & 94.51$_{\pm0.11}$ & 18.03$_{\pm0.28}$ \\
        SGC  & 97.55$_{\pm0.57}$ & 97.76$_{\pm0.44}$ & 95.76$_{\pm0.13}$ & 96.21$_{\pm0.09}$ & 92.18$_{\pm0.32}$ & 95.50$_{\pm0.16}$ & 94.02$_{\pm0.03}$ & 92.40$_{\pm0.09}$ & 16.50$_{\pm0.91}$ \\
        \rowcolor{gray!20}
        Average & 97.48 & 97.72 & 96.74 & 97.05 & 93.28 & 95.80 & 94.17 & 92.91 & 17.24 \\
        \midrule
        GC-SNN$^*$ & 95.14$_{\pm0.59}$ & 95.85$_{\pm0.50}$ & 91.00$_{\pm0.37}$ & 91.87$_{\pm0.26}$ & \second{94.48$_{\pm0.01}$} & 92.87$_{\pm0.02}$ & 84.36$_{\pm0.51}$ & 91.29$_{\pm0.25}$ & \second{8.54$_{\pm0.38}$} \\
        GA-SNN$^*$ & 91.84$_{\pm1.68}$ & 93.35$_{\pm1.34}$ & 87.59$_{\pm0.20}$ & 88.29$_{\pm0.22}$ & OOM & OOM & 92.72$_{\pm0.33}$ & 91.46$_{\pm0.36}$ & OOM \\
        SpikingGCN & 97.35$_{\pm0.44}$ & 97.28$_{\pm0.36}$ & 96.47$_{\pm0.14}$ & 96.82$_{\pm0.09}$ & 92.07$_{\pm0.38}$ & \third{95.32$_{\pm0.25}$} & 92.35$_{\pm0.36}$ & 91.06$_{\pm0.42}$ & \first{11.68$_{\pm0.41}$} \\
        SpikeNet   & 94.15$_{\pm1.44}$ & 95.38$_{\pm0.86}$ & 93.52$_{\pm0.84}$ & 94.59$_{\pm0.60}$ & OOM & OOM & 93.03$_{\pm0.07}$ & 91.78$_{\pm0.05}$ & OOM \\
        SiGNN      & 95.21$_{\pm0.58}$ & 96.05$_{\pm0.39}$ & 95.30$_{\pm0.32}$ & 96.08$_{\pm0.21}$ & OOM & OOM & 90.96$_{\pm0.35}$ & 90.40$_{\pm0.49}$ & OOM \\
        SpikeGCL   & \first{99.30$_{\pm0.11}$} & \first{99.17$_{\pm0.14}$} & \first{99.04$_{\pm0.07}$} & \first{99.03$_{\pm0.06}$} & \first{94.72$_{\pm0.04}$} & \first{96.56$_{\pm0.03}$} & \first{95.07$_{\pm0.21}$} & \first{93.35$_{\pm0.27}$} & OOM \\
        SpikeGT    & \third{97.80$_{\pm0.50}$} & \second{97.98$_{\pm0.34}$} & \third{96.53$_{\pm0.14}$} & \third{96.97$_{\pm0.09}$} & 85.50$_{\pm0.59}$ & 92.93$_{\pm0.40}$ & \second{94.36$_{\pm0.02}$} & \second{92.76$_{\pm0.02}$} & OOM \\
        MSG        & \second{97.93$_{\pm0.38}$} & \third{97.59$_{\pm0.33}$} & \second{97.74$_{\pm0.11}$} & \second{97.74$_{\pm0.11}$} & \third{93.54$_{\pm0.20}$} & \second{95.90$_{\pm0.15}$} & \third{94.35$_{\pm0.46}$} & \third{92.67$_{\pm0.42}$} & \third{4.13$_{\pm1.14}$} \\
        DRSGNN     & 96.37$_{\pm0.25}$ & 95.45$_{\pm0.30}$ & 85.28$_{\pm1.37}$ & 87.20$_{\pm0.51}$ & >3days & >3days & 75.53$_{\pm0.85}$ & 75.16$_{\pm0.59}$ & >3days \\
        \rowcolor{gray!20}
        Average    & 96.12 & 96.46 & 93.61 & 94.29 & 92.06 & 94.72 & 90.30 & 89.99 & 8.12 \\
        \bottomrule
    \end{tabular}
    \vspace{-10px}
\end{table}

\begin{table}[t]
    \centering
    \scriptsize
    \caption{Performance comparison on DBLP, Tmall and Patent datasets.}
    \label{tab:dyn_nc}
    \resizebox{\textwidth}{!}{
    \begin{tabular}{l|cc|cc|cc}
        \toprule
        \multirow{2}{*}{\textbf{Models}} & \multicolumn{2}{c|}{\textbf{DBLP}} & \multicolumn{2}{c|}{\textbf{Tmall}} & \multicolumn{2}{c}{\textbf{Patent}} \\
        \cmidrule(lr){2-3} \cmidrule(lr){4-5} \cmidrule(lr){6-7}
        & \textbf{Macro-F1} & \textbf{Micro-F1} & \textbf{Macro-F1} & \textbf{Micro-F1} & \textbf{Macro-F1} & \textbf{Micro-F1} \\
        \midrule
        ROLAND   & 69.92$_{\pm0.91}$ & 70.49$_{\pm0.55}$ & 52.69$_{\pm0.67}$ & 59.08$_{\pm0.52}$ & OOM & OOM \\
        ScaDyG   & 70.76$_{\pm0.87}$ & 71.98$_{\pm0.24}$ & 55.05$_{\pm0.49}$ & 60.28$_{\pm0.49}$ & OOM & OOM \\
        GraphSSM & 70.86$_{\pm0.14}$ & 71.68$_{\pm0.11}$ & 57.57$_{\pm0.22}$ & 62.08$_{\pm0.07}$ & OOM & OOM \\
        \rowcolor{gray!20}
        Average  & 70.51 & 71.38 & 55.10 & 60.48 & - & - \\
        \midrule
        TGAT      & 70.38$_{\pm0.30}$ & 71.50$_{\pm0.09}$ & 55.03$_{\pm0.15}$ & 60.59$_{\pm0.12}$ & 84.90$_{\pm0.03}$ & 84.53$_{\pm0.06}$ \\
        DyGFormer & 68.13$_{\pm0.41}$ & 69.70$_{\pm0.22}$ & 52.22$_{\pm0.51}$ & 59.63$_{\pm0.10}$ & 85.47$_{\pm0.04}$ & 85.23$_{\pm0.10}$ \\
        TPNet     & 68.13$_{\pm0.50}$ & 68.95$_{\pm0.39}$ & 43.59$_{\pm0.38}$ & 48.05$_{\pm0.03}$ & OOM & OOM \\
        \rowcolor{gray!20}
        Average   & 68.88 & 70.05 & 50.28 & 56.09 & 85.19 & 84.88 \\
        \midrule
        GC-SNN
        & 70.70$_{\pm0.51}$ & 71.73$_{\pm0.48}$ & 58.27$_{\pm0.13}$ & 62.60$_{\pm0.25}$ & 82.62$_{\pm0.78}$ & 82.52$_{\pm0.78}$ \\
        SpikingGCN & 68.70$_{\pm0.48}$ & 69.55$_{\pm0.37}$ & 55.49$_{\pm0.26}$ & 60.92$_{\pm0.23}$ & 78.66$_{\pm0.34}$ & 78.60$_{\pm0.81}$ \\
        SpikeGT    & 68.58$_{\pm0.38}$ & 69.49$_{\pm0.65}$ & 56.98$_{\pm0.27}$ & 61.26$_{\pm0.42}$ & 81.50$_{\pm0.85}$ & 81.37$_{\pm0.97}$ \\
        SpikeNet   & \third{71.38$_{\pm0.47}$} & \third{72.03$_{\pm0.29}$} & \third{59.07$_{\pm0.59}$} & \third{63.42$_{\pm0.60}$} & \third{83.92$_{\pm0.85}$} & \third{83.89$_{\pm0.73}$} \\
        SiGNN      & \second{73.74$_{\pm0.29}$} & \second{74.66$_{\pm0.65}$} & \first{61.87$_{\pm0.21}$} & \first{65.74$_{\pm0.15}$} & \second{84.10$_{\pm0.76}$} & \second{84.06$_{\pm0.51}$} \\
        HASNN      & \first{75.23$_{\pm0.28}$} & \first{75.41$_{\pm0.25}$} & \second{59.34$_{\pm0.34}$} & \second{64.05$_{\pm0.09}$} & \first{84.35$_{\pm0.71}$} & \first{84.31$_{\pm0.62}$} \\
        \rowcolor{gray!20}
        Average    & 71.39 & 72.15 & 58.50 & 63.00 & 82.53 & 82.46 \\
        \bottomrule
    \end{tabular}
    }
    \vspace{-10px}
\end{table}

\subsection{Comparison on Dynamic Graphs}
In this section, we evaluate SGNNs and classic full-precision representation learning methods on large-scale dynamic graphs, as reported in Table~\ref{tab:dyn_nc}. 
\textbf{On DBLP and Tmall, SGNNs outperform discrete-time and continuous-time GNNs by an average of 2.94\% and 8.05\%, respectively, across the multiple evaluation metrics.} These results highlight the potential of SGNNs as promising energy-efficient alternatives to conventional discrete-time GNNs. Compared with continuous-time methods, SGNNs achieve competitive or superior performance on DBLP and Tmall but still fall behind continuous-time GNNs on Patent. Most existing SGNNs process dynamic graphs as discrete snapshots with coarse-grained timestamps, which may limit their ability to capture fine-grained temporal interactions. Consequently, modeling continuous-time graphs without node features remains challenging for existing SGNNs. We provide further discussion in Appendix~\ref{sup:extra_exp_dyn}.

In addition, some SGNNs are built upon the multi-step mode, where single-step or multi-step modes correspond to step-by-step or layer-by-layer propagation strategies \cite{fang2023spikingjelly}. Although the multi-step mode facilitates matrix/tensor operations by merging time steps into batch dimensions, storing multi-step intermediate node embeddings incurs a non-negligible memory overhead for large-scale graphs. This mode makes it challenging for SGNNs to scale to dynamic graphs with longer temporal durations, such as Tmall (186 timesteps). Besides, another bottleneck is that some neural network components tailored to the multi-step mode rely on the outputs from all previous steps. This severely limits the scalability of SGNNs when dealing with dynamically evolving graph data.

\begin{observation}
    For dynamic graphs, state-of-the-art SGNNs demonstrate highly competitive performance and have outperformed classic full-precision graph learning methods in node classification tasks. However, \textbf{balancing data parallelism and scalability of SGNNs is still a crucial direction for further exploration.}
\end{observation}

\begin{table}[tbp]
    \centering
    \caption{The model size (MB), maximum memory usage (GB) and theoretical energy consumption (mJ) of full-precision GNNs and SGNNs.} \label{tab:energy_analysis}
    \resizebox{\textwidth}{!}{
    \begin{tabular}{l|ccc|ccc|ccc|ccc}
        \toprule
        \multirow{2}*{\textbf{Models}} & \multicolumn{3}{c|}{\textbf{PubMed}} & \multicolumn{3}{c|}{\textbf{Computers}} & \multicolumn{3}{c|}{\textbf{Physics}} & \multicolumn{3}{c}{\textbf{ogbn-arxiv}} \\
        \cmidrule(lr){2-4} \cmidrule(lr){5-7} \cmidrule(lr){8-10} \cmidrule(lr){11-13}
        & \textbf{Size$\downarrow$} & \textbf{Mem$\downarrow$} & \textbf{Eng$\downarrow$} & \textbf{Size$\downarrow$} & \textbf{Mem$\downarrow$} & \textbf{Eng$\downarrow$} & \textbf{Size$\downarrow$} & \textbf{Mem$\downarrow$} & \textbf{Eng$\downarrow$} & \textbf{Size$\downarrow$} & \textbf{Mem$\downarrow$} & \textbf{Eng$\downarrow$} \\
        \midrule
        GCN        & 0.74 & 0.55 & 23.42 & 1.13 & 1.15 & 25.35 & 12.33 & 2.02  & 648.26  & 0.21 & 3.17  & 60.47 \\
        SAGE       & 0.98 & 0.65 & 23.56 & 1.51 & 1.91 & 26.79 & 16.44 & 20.23 & 703.14  & 0.27 & 1.73  & 60.38 \\
        GAT        & 1.47 & 1.14 & 58.84 & 2.26 & 4.26 & 64.80 & 24.66 & 5.45  & 1712.11 & 0.40 & 19.10 & 143.75 \\      
        SGC        & 0.74 & 0.49 & 17.66 & 1.13 & 0.52 & 19.42 & 12.33 & 4.69  & 513.62  & 0.21 & 0.99  & 48.30 \\ 
        \midrule 
        GC-SNN     & 0.49 & 2.90  & 4.60 & 0.76 & 2.29 & 5.00 & 4.18 & 10.05 & 133.93 & 0.15 & 22.43 & 10.86 \\
        GA-SNN     & 0.56 & 3.46 & 9.25 & 0.82 & 8.04 & 10.23 & 8.29  & 11.18 & 268.10  & -    & -     & - \\    
        SpikingGCN & 0.25 & \third{0.64} & \first{0.49}   & 0.38 & \second{1.19} & \first{0.57}   & 4.11  & \first{2.28}  & \first{13.48}   & 0.08 & \third{4.26}  & \first{4.22} \\
        SpikeNet   & 0.57 & \second{0.53} & 5.24  & 0.86 & \third{1.72} & 5.40  & 8.31  & 17.40 & 137.44  & 0.38 & \second{2.99}  & 16.82 \\
        SpikeGCL   & \third{0.17} & 1.81 & \second{0.52} & \third{0.24} & 2.38 & \second{0.68} & \third{2.22}  & 8.78  & \second{13.70}   & \third{0.07} & 15.25 & \second{4.59} \\
        SpikeGT    & 1.80 & 5.56 & 28.74  & 1.96 & 4.17 & 22.60  & 5.82  & 14.48 & 189.01  & 1.67 & 22.26 & 229.23 \\
        DRSGNN     & \first{0.01} & \first{0.41} & \third{1.49}   & \first{0.03} & \first{1.08} & 5.76   & \first{0.16}  & \third{8.12}  & 24.27   & \first{0.02} & \first{0.59}  & 148.53 \\
        MSG        & \second{0.07} & 2.01 & 1.66   & \second{0.11} & 4.17 & \third{1.44}   & \second{1.07}  & \second{5.85}  & \third{15.57}   & \second{0.03} & 20.19 & \third{11.71} \\
        SiGNN      & 3.13 & 0.67 & 52.19 & 4.70 & 3.97 & 55.78 & 49.70 & 18.62 & 1532.00 & 1.16 & 7.28  & 115.96 \\
        \bottomrule
    \end{tabular}
    }
\end{table}

\section{Benchmarking Over Efficiency (RQ2)}\label{sec:exp_efficiency}
\subsection{Experimental Setup for Energy Efficiency Analysis}
There is a widely used analytical approach that evaluates the energy consumption of SGNNs by counting the overall peak activity during inference \cite{zhu2022spiking}. However, we argue that this oversimplified approach significantly underestimates the energy cost of SGNNs. For example, a computationally intensive SGNN may be wrongly evaluated as an energy-efficient model by indiscriminately applying activity regularization to suppress excessive spiking. Therefore, we follow the theoretical energy consumption formula proposed in \cite{yao2024spike}. The energy cost of SGNNs is computed by summing the per-layer contributions based on the layer's theoretical FLOPs and the corresponding spike firing rate:
\begin{align}
\textbf{SNNs:} & \; E_{snn}=\sum_{l}E_{ac}\times FP_{l}\times T \times R_{l}, \\
\textbf{ANNs:} & \; E_{ann}=\sum_{l}E_{mac}\times FP_{l},
\end{align}
where $E_{mac}$ and $E_{ac}$, the energy of multiply-accumulate (MAC) and accumulate (AC) operations, are set to $4.5pJ$ and $0.9pJ$ \cite{zhou2023spikingformer}. $FP_{l}$ and $R_l$ denote the number of floating-point operations and spiking firing rate of the $l$-th layer. Appendix~\ref{sup:eng_consumption} further provides detailed calculations of theoretical energy consumption for different models and additional energy efficiency analyses on dynamic graphs.

\subsection{Energy Efficiency Analysis}\label{sec:6_2}
The complete energy efficiency analysis results are presented in Table~\ref{tab:energy_analysis}. GNNs typically have a small number of parameters but a high computational workload. Integrating the lightweight, low-power LIF models into GNNs does not significantly increase the overall model size. Except for SpikeGT and SiGNN, the other SGNNs often have similar or even smaller model sizes across the four datasets, contributing to the edge deployment of SGNNs.
As shown in Table~\ref{tab:energy_analysis}, the methods built upon the single-step mode (SpikingGCN and DRSGNN) have smaller memory footprints than the multi-step approaches (GC-SNN, GA-SNN and SpikeGT). 
SpikeNet and SiGNN are more sensitive to the dimensionality of node features, which leads to increased energy consumption as the feature dimensionality grows. Besides, we believe specialized neuromorphic hardware like Loihi offers potential solutions to mitigate this issue concerning memory usage \cite{davies2018loihi}.

\begin{observation}
    While some widely accepted SNN practices in other domains may backfire when transferred to graph data, careful architectural design ensures that incorporating spiking neurons does not necessarily lead to larger model size or higher training memory usage.
\end{observation}

In addition, the theoretical energy consumption is another key consideration, especially for large-scale graph applications. SpikingGCN and SpikeGCL have the lowest and second-lowest energy consumption on all datasets, respectively. \textbf{Compared to the full-precision SGC, SpikingGCN and SpikeGCL respectively achieve 29.91× and 27.63× less average energy consumption across the four datasets.} Furthermore, GC-SNN and GA-SNN achieve 5.14× and 6.36× lower energy consumption compared to their full-precision counterparts GCN and GAT, respectively. 
DRSGNN incorporates random walk–based local encoding, which significantly increases computational overhead. Besides, the experimental results in Table~\ref{tab:energy_analysis} reveal some undesired architectural designs of SGNNs. SiGNN retains the computed results from all previous timesteps and repeatedly deploys aggregators to compute node embeddings at each timestep. Although this design indeed achieves better predictive performance, its energy consumption exceeds that of most full-precision GNNs on four datasets.

\begin{observation}
    Replacing artificial neurons with spiking neurons does offer a viable pathway toward constructing energy-efficient neural networks for graph-structured data. The main energy bottleneck in most SGNNs stems from their GNN layers, which are originally designed for high-precision vectors and fail to fully exploit the computation-friendly nature of binary spiking representations.
\end{observation}

\begin{table}[tbp]
    \centering
    \scriptsize
    % \tabcolsep=2px
    \caption{Ablation studies on homophilic and heterophilic graphs.}\label{tab:ablation}
    \begin{tabular}{l|c|cccccc}
        \toprule
        \textbf{Models} & \textbf{Ablation} &\textbf{PubMed} & \textbf{Computers} & \textbf{CS} & \textbf{Squirrel} & \textbf{Roman-Empire} & \textbf{Minesweeper} \\
        \midrule
        \multirow{5}{*}{GC-SNN} & Vanilla & 77.35$_{\pm0.65}$ & 91.03$_{\pm0.88}$ & 93.68$_{\pm1.71}$ & \first{43.49$_{\pm1.30}$} & 55.79$_{\pm0.57}$ & 82.27$_{\pm0.70}$ \\
        & STFN $\rightarrow$ BN & 74.20$_{\pm1.85}$ & 90.59$_{\pm0.44}$ & 91.12$_{\pm0.37}$ & 39.54$_{\pm1.32}$ & 52.95$_{\pm0.38}$ & 82.11$_{\pm0.73}$ \\
        & LIF $\rightarrow$ PLIF & 78.22$_{\pm0.65}$ & 91.65$_{\pm0.75}$ & 93.65$_{\pm1.55}$ & 43.33$_{\pm1.91}$ & 55.14$_{\pm1.18}$ & 82.38$_{\pm0.80}$ \\
        & add MS & 77.40$_{\pm0.70}$ & \first{93.06$_{\pm3.00}$} & \first{97.10$_{\pm1.58}$} & 43.19$_{\pm0.70}$ & 57.32$_{\pm0.06}$ & 81.80$_{\pm0.66}$ \\
        & add JK & \first{78.83$_{\pm0.70}$} & 91.54$_{\pm0.40}$ & 94.08$_{\pm1.47}$ & 43.00$_{\pm1.25}$ & \first{63.62$_{\pm0.52}$} & \first{85.46$_{\pm0.59}$} \\
        \midrule
        \multirow{5}{*}{GA-SNN} & Vanilla & 76.10$_{\pm2.16}$ & 88.15$_{\pm0.71}$ & 90.31$_{\pm0.41}$ & 40.45$_{\pm0.82}$ & 65.91$_{\pm1.02}$ & 87.92$_{\pm0.27}$ \\
        & STFN $\rightarrow$ BN & 70.00$_{\pm4.09}$ & 88.82$_{\pm1.15}$ & 89.66$_{\pm1.02}$ & \first{42.45$_{\pm1.56}$} & 68.84$_{\pm0.11}$ & 88.10$_{\pm0.42}$ \\
        & LIF $\rightarrow$ PLIF & 74.22$_{\pm2.62}$ & 88.06$_{\pm1.48}$ & 90.41$_{\pm0.28}$ & 39.62$_{\pm0.56}$ & 68.92$_{\pm0.50}$ & 87.75$_{\pm0.81}$ \\
        & add MS & \first{77.42$_{\pm1.99}$} & \first{90.47$_{\pm1.13}$} & \first{95.11$_{\pm0.55}$} & 41.78$_{\pm2.03}$ & \first{74.06$_{\pm0.82}$} & \first{90.36$_{\pm0.37}$} \\
        & add JK & 75.04$_{\pm2.32}$ & 88.35$_{\pm0.92}$ & 92.17$_{\pm1.95}$ & 40.75$_{\pm0.98}$ & 68.00$_{\pm0.70}$ & 88.68$_{\pm0.40}$ \\
        \midrule
        \multirow{5}{*}{SpikingGCN} & Vanilla & 74.03$_{\pm1.18}$ & 85.64$_{\pm1.01}$ & 91.71$_{\pm0.24}$ & 41.49$_{\pm0.87}$ & 65.23$_{\pm1.04}$ & 74.17$_{\pm1.13}$ \\
        & BN $\rightarrow$ STFN  & \first{77.14$_{\pm2.29}$} & 88.06$_{\pm0.68}$ & 92.89$_{\pm0.14}$ & 41.62$_{\pm1.24}$ & 61.74$_{\pm0.20}$ & 74.08$_{\pm0.96}$ \\
        & LIF $\rightarrow$ PLIF & 76.40$_{\pm1.85}$ & 91.30$_{\pm0.15}$ & 93.36$_{\pm0.48}$ & 43.27$_{\pm0.28}$ & 67.72$_{\pm1.05}$ & 74.32$_{\pm0.95}$ \\
        & add MS & 76.90$_{\pm1.28}$ & \first{91.75$_{\pm0.52}$} & 95.25$_{\pm0.29}$ & \first{43.79$_{\pm0.18}$} & \first{72.09$_{\pm0.62}$} & \first{90.76$_{\pm0.72}$} \\
        & add JK & 75.44$_{\pm2.42}$ & 91.13$_{\pm0.40}$ & \first{95.78$_{\pm0.17}$} & 43.12$_{\pm0.96}$ & 65.92$_{\pm0.80}$ & 75.96$_{\pm1.16}$ \\
        \bottomrule
    \end{tabular}
\end{table}

\section{Ablation Study (RQ3)}\label{sec:exp_ablation}
In this section, we conducted ablation studies to investigate the effectiveness of classic neural network components and spiking neuron variants in SGNNs. GC-SNN, GA-SNN and SpikingGCN are selected as representative baselines, whose network backbones can be considered as simplified GCN, GAT and SGC, respectively. We evaluate the impact of individual components on both homophilic and heterophilic graphs by replacing BatchNorm \cite{ioffe2015batch} with STFNorm \cite{xu2021exploiting}, replacing LIF \cite{gerstner2014neuronal} with PLIF \cite{fang2021incorporating}, and incorporating Membrane Skip connections \cite{yao2024spike} (MS) and jumping knowledge \cite{xu2018representation} (JK). We provide design specifications and significance tests in Appendix~\ref{sup:extra_ablation}.

The experimental results are demonstrated in Table~\ref{tab:ablation}. Overall, MS significantly boosts the performance of baselines on most datasets. Specifically, across multiple homophilic graphs, GC-SNN, GA-SNN and SpikingGCN achieve average accuracy improvements of 1.83, 2.81 and 4.17 percentage points, respectively. Besides, MS provides consistent performance gains for both GA-SNN and SpikingGCN on heterophilic datasets. SpikingGCN with MS achieves an AUC improvement of 16.59 percentage points on Minesweeper. 
Notably, for GC-SNN, JK leads to improvements of 7.83 and 3.19 percentage points on the Roman-Empire and Minesweeper datasets. MS and JK serve as effective approaches to mitigate the performance degradation caused by deeper architectures of SGNNs.
In addition, the advanced spiking neuron with learnable parameters, PLIF, enables SpikingGCN to achieve better performance across all datasets. However, the performance gains brought by PLIF are not significant for GC-SNN and GA-SNN. 
STFNorm, which is designed to accommodate the dynamics of spiking neurons, results in noticeable performance improvements on homophilic graphs. STFNorm leads to consistent improvements only for GC-SNN, with gains of 3.95, 2.84 and 0.16 percentage points across three heterophilic graphs. 

\begin{observation}
    Neural network components in traditional GNNs, like skip connections and jumping knowledge, are similarly beneficial for SGNNs. The effectiveness of spiking neuron and normalization variants varies significantly, indicating a lack of generalizable improvement across different graphs.
\end{observation}
 
\section{Conclusion}
In this paper, we propose a fair and consistent benchmark, \textbf{SGNNBench}, to evaluate the effectiveness and efficiency of various SGNN branches. We conduct an extensive and in-depth comparative study, analyzing various SGNNs and evaluating the strengths and weaknesses of different SGNN network structures. Through energy consumption analysis, we highlight energy bottlenecks in SGNNs to promote rigorous empirical standards in the field.
We hope that this comprehensive study of benchmarks and reflections establishes a solid, practical and systematic foundation for the SGNN community, providing future researchers with deeper insights into SNNs and energy-efficient GNNs.

\begin{ack}
This work was supported in part by the Guangdong Basic and Applied Basic Research Foundation, China under Grant 2023A1515011050, in part by Shenzhen Science and Technology Program under Grant KJZD20231023094501003, and in part by Tencent AI Lab under Grant RBFR2024004.
\end{ack}

\bibliographystyle{abbrv}
\bibliography{reference}

%%%%%%%%%%%%%%%%%%%%%%%%%%%%%%%%%%%%%%%%%%%%%%%%%%%%%%%%%%%%
\clearpage
\appendix

\section{Experimental Setting}\label{sup:exp_setting}

\subsection{Datasets}
In Table~\ref{tab:dataset_stat}, we summarize the characteristics and statistics of 24 datasets. For three citation networks \cite{sen2008collective} (Cora, CiteSeer, PubMed), the same split strategy as in the previous semi-supervised setting \cite{kipf2016semi} is adopted. For two co-purchase networks \cite{shchur2018pitfalls} (Computers and Photo) and two co-author networks (CS and Physics), we follow the widely accepted training/validation/test split of 60\%/20\%/20\%. For Squirrel and Chameleon, we adopt a new data split proposed in \cite{platonov2023critical} that removes overlapping nodes.
In addition, for six OGB datasets \cite{hu2020open} (ogbn-proteins, ogbn-arxiv, ogbn-products, ogbl-collab, ogbl-ddi and ogbl-ppa) and four heterophilic datasets proposed in \cite{platonov2023critical} (Roman-Empire, Amazon-Ratings, Minesweeper and Questions), we utilize their official splits.

For dynamic graph learning, we include three real-world datasets (DBLP \cite{lu2019temporal}, Tmall \cite{lu2019temporal} and Patent \cite{NBERw8498}) for node classification. Following \cite{li2023scaling}, we adopt a 40\% training ratio and report Macro-F1 and Micro-F1 scores. We further introduce two temporal graph datasets from the Temporal Graph Benchmark (tgbl-wiki and tgbl-review) for dynamic link prediction \cite{gastinger2024tgb}. Following the common practice~\footnote{\url{https://github.com/fpour/TGB_Baselines}}, we apply temporal bucketing to aggregate timestamped interactions into graph snapshots, enabling discrete-time GNNs and SGNNs to process these continuous-time datasets.

\begin{table}[htbp]
    \centering
    \small
    \tabcolsep=3.5px
    \caption{Overview of dataset statistics. $T$ is the total number of time steps in dynamic graphs.}\label{tab:dataset_stat}
    \begin{tabular}{l|c|ccc}
    \toprule
    \textbf{Dataset} & \textbf{Type} & \textbf{\#nodes} & \textbf{\#edges} & \textbf{Metric}\\
    \midrule
    Cora      & Homophily, Static & 2,708  & 10,556  & Accuracy \\
    CiteSeer  & Homophily, Static & 3,327  & 9,104   & Accuracy \\
    PubMed    & Homophily, Static & 19,717 & 88,684  & Accuracy \\
    Computers & Homophily, Static & 13,752 & 491,722 & Accuracy \\
    Photo     & Homophily, Static & 7,650  & 238,162 & Accuracy \\
    CS        & Homophily, Static & 18,333 & 163,788 & Accuracy \\
    Physics   & Homophily, Static & 34,493 & 495,924 & Accuracy \\
    \midrule
    ogbn-proteins & Homophily, Static & 132,534   & 39,561,252 & ROC-AUC \\
    ogbn-arxiv    & Homophily, Static & 169,343   & 1,166,243  & Accuracy \\
    ogbn-products & Homophily, Static & 2,449,029 & 61,859,140 & Accuracy \\
    \midrule
    ogbl-collab   & Homophily, Static & 235,868   & 1,285,465  & AUC\&AUPR \\
    ogbl-ddi      & Homophily, Static & 4,267     & 1,334,889  & AUC\&AUPR \\ 
    ogbl-ppa      & Homophily, Static & 576,289   & 30,326,273 & Hits@100 \\
    \midrule
    Squirrel       & Heterophily, Static & 2,223  & 46,998  & Accuracy \\
    Chameleon      & Heterophily, Static & 890    & 8,854   & Accuracy \\
    Roman-Empire   & Heterophily, Static & 22,662 & 32,927  & Accuracy \\
    Amazon-Ratings & Heterophily, Static & 24,492 & 93,050  & Accuracy \\
    Minesweeper    & Heterophily, Static & 10,000 & 39,402  & ROC-AUC \\
    Questions      & Heterophily, Static & 48,921 & 153,540 & ROC-AUC \\
    \midrule
    DBLP   & Dynamic, $T=27$  & 28,085    & 236,894    & Macro\&Micro-F1 \\
    Tmall  & Dynamic, $T=186$ & 577,314   & 4,807,545  & Macro\&Micro-F1 \\
    Patent & Dynamic, $T=25$  & 2,738,012 & 13,960,811 & Macro\&Micro-F1 \\
    \midrule
    tgbl-wiki   & Dynamic, $T=152,757$ & 9,227   & 157,474   & MRR \\
    tgbl-review & Dynamic, $T=6,865$   & 352,637 & 4,873,540 & MRR \\
    \bottomrule
    \end{tabular}
\end{table}

\subsection{Baselines}
We collect representative SGNNs and full-precision GNNs for a comprehensive comparison across static and dynamic graph learning tasks. The evaluated baselines are organized as follows:

\begin{itemize}
    \item \textbf{Spiking Graph Neural Networks.}
    We implement 10 SGNN baselines, including GC-SNN and GA-SNN \cite{xu2021exploiting}, SpikingGCN \cite{zhu2022spiking}, SpikeNet \cite{li2023scaling}, SpikeGCL \cite{li2023graph}, SpikeGT \cite{sun2024spikegraphormer}, MSG \cite{sun2024spiking}, DRSGNN \cite{zhao2024dynamic}, SiGNN \cite{chen2025signn} and HASNN \cite{gan2026hasnn}.
    
    \item \textbf{Full-precision GNNs on Static Graphs.}
    We consider four classic GNNs: GCN \cite{kipf2016semi}, SAGE \cite{hamilton2017inductive}, GAT \cite{velivckovic2017graph} and SGC \cite{wu2019simplifying}. Following recent studies \cite{sancak2024we, luo2024classic}, we incorporate normalization, dropout and skip connections into these classic GNNs for fair and rigorous comparisons.
    We additionally include two advanced GNNs (GPRGNN \cite{chien2021gprgnn} and ChebNet\MakeUppercase{\romannumeral 2} \cite{he2022convolutional}) and two Graph Transformers (SGFormer \cite{wu2023sgformer} and Polynormer \cite{deng2024polynormer}) on homophilic graphs. For heterophilic graphs, we adopt heterophily-robust GNNs: H2GCN \cite{zhu2020beyond} and FSGNN \cite{maurya2022simplifying}.
    
    \item \textbf{Full-precision GNNs on Dynamic Graphs.}
    We evaluate three discrete-time methods (ROLAND \cite{you2022roland}, ScaDyG \cite{wu2026scadyg} and GraphSSM \cite{li2024state}) and three continuous-time methods (TGAT \cite{xu2020inductive}, DyGFormer \cite{yu2023towards} and TPNet \cite{lu2024improving}).
\end{itemize}

Except for SpikeNet, MSG and SiGNN, which design specialized neurons for their networks, other SGNNs utilize the LIF model as their default spiking neuron. For GC-SNN and GA-SNN, we provide our unofficial implementation due to the unavailability of the source code. We consistently adopt the neighbor sampling strategy proposed in \cite{li2023scaling} for all SGNN baselines on dynamic graphs.

\subsection{Hyperparameter Setting and Reproducibility}
To establish a rigorous and comprehensive benchmark, we adopt the Tree-structured Parzen Estimator Approach \cite{bergstra2011algorithms} for hyperparameter tuning. The hyperparameter search spaces are defined based on the default configurations specified by the authors in their original implementations. We present the corresponding hyperparameter search spaces in Table~\ref{tab:8}. 
Specifically, for SpikeNet and SiGNN, we adhere to their default neighbor sampling strategy, which samples a fixed number of reachable neighbors within 2 hops. Except for SpikeGT and DRSGNN, all other SGNNs follow the same search space of time steps in the general settings.

\begin{table}[htbp]
    \centering
    \small
    \tabcolsep=3.0px
    \caption{Hyperparameter search space of SGNNs.}\label{tab:8}
    \begin{tabular}{l|cccc}
    \toprule
    \textbf{Models} & \textbf{Hyperparameter} & \textbf{Values} \\
    \midrule
    \multirow{9}{*}{\makecell{General \\ Settings}} & Learning Rate & [0.1, 0.01, 0.001, 0.0001] \\
    & Weight Decay    & [0.0, 0.1, 0.01, 0.001, 0.0001] \\
    & Epochs          & [100, 200, 500, 1000] \\
    & Hidden Channels & [32, 64, 128, 256, 512] \\
    & GNN Layers      & [1, 2, 4, 6, 8, 10, 12] \\
    & Heads           & [1, 2, 4] \\
    & Dropout Rate    & [0.0, 0.2, 0.4, 0.6] \\
    & $U_{th}$        & [1.0, 0.5, 0.1] \\
    & $T$             & [2, 5, 10, 15, 20] \\
    \midrule
    SpikingGCN & Hops & [1, 2, 4, 6, 8] \\
    \midrule
    \multirow{3}{*}{\makecell{SpikeNet \\ SiGNN \\ HASNN}}   & Neighbors & [[5, 2], [8, 5], [10, 8]] \\
    & Hidden Channels & [[128, 64], [256, 128], [512, 256]] \\
    & Batch Size      & [1024, 2048, 4096, 8192] \\ 
    \midrule
    SpikeGCL   & Drop Edge Rate & [0.2, 0.4, 0.6, 0.8]  \\
    \midrule
    \multirow{3}{*}{SpikeGT}    & Transformer Layers & [1, 2, 3, 4] \\
    & GNN Branch Weights & [0.1, 0.3, 0.5, 0.7, 0.9] \\
    & $T$ & [2, 4, 6, 8, 10] \\
    \midrule
    \multirow{2}{*}{MSG} & Backbone & [euclidean, sphere, lorentz] \\
    & Step Size & [1.0, 0.1, 0.01] \\
    \midrule
    \multirow{2}{*}{DRSGNN}     & Walk Length & [8, 16, 32] \\
    & $T$ & [100, 200, 300, 400, 500] \\
    \bottomrule
    \end{tabular}
\end{table}

\subsection{Computation Resources}\label{sup:computational_resource}
All experiments are conducted on two RTX 3090Ti GPUs and an AMD Ryzen Threadripper processor. In baseline comparisons, experiment settings that exceed the GPU memory limit are reported as out-of-memory (OOM). Besides, we track the elapsed time of each experiment, and any case that exceeds three days is reported as >3days.

\section{Additional Results on Large-scale Graphs}\label{sup:extra_exp_static}

Table~\ref{tab:products_nc} reports the node classification results on ogbn-products, a large-scale graph containing over 2.4 million nodes and 61 million edges. \textbf{Overall, SGNNs achieve approximately 97\% of the average performance of full-precision GNNs.} Notably, SpikingGCN and MSG outperform several full-precision baselines, demonstrating promising performance despite the extreme quantization introduced by binary spiking representations. 
However, the scalability of existing SGNNs remains constrained by their architectural designs. Several methods (e.g., SpikeNet, SiGNN and SpikeGCL) encounter out-of-memory errors, potentially due to the substantial memory overhead of storing intermediate spiking representations across multiple time steps. Moreover, graph preprocessing operations, such as structural and positional encoding, often lack efficient spike-driven alternatives and can incur substantial computational overhead on large-scale graphs. 
These findings suggest that there is no silver bullet for achieving both efficiency and scalability in SGNNs. Specialized GNN architectures are needed to fully exploit the computational advantages of binary spikes.

\begin{table}[htbp]
    \centering
    \caption{Node classification on ogbn-products in terms of accuracy (\%).}
    \label{tab:products_nc}
    \begin{tabular}{l|c}
        \toprule
        \textbf{Models} & \textbf{ACC (\%)} \\
        \midrule
        GCN        & 63.73$_{\pm0.23}$ \\
        SAGE       & 62.03$_{\pm0.22}$ \\
        GAT        & 64.30$_{\pm0.10}$ \\
        SGC        & 61.96$_{\pm0.01}$ \\
        \rowcolor{gray!20}
        Average    & 63.01 \\
        \midrule
        GC-SNN     & 59.25$_{\pm0.04}$ \\
        GA-SNN     & 58.69$_{\pm0.01}$ \\
        SpikingGCN & \first{63.74$_{\pm0.21}$} \\
        SpikeNet   & OOM \\
        SiGNN      & OOM \\
        SpikeGCL   & OOM \\
        SpikeGT    & \third{61.51$_{\pm0.10}$} \\
        MSG        & \second{62.47$_{\pm0.29}$} \\
        DRSGNN     & >3days \\
        \rowcolor{gray!20}
        Average    & 61.13 \\
        \bottomrule
    \end{tabular}
\end{table}

\section{Additional Analyses on Dynamic Graphs}\label{sup:extra_exp_dyn}

\subsection{Performance on Temporal Link Prediction}
Table~\ref{tab:tgbl} reports the temporal link prediction results on tgbl-wiki and tgbl-review in terms of Mean Reciprocal Rank (MRR). \textbf{To the best of our knowledge, this is the first evaluation of SGNNs under the Temporal Graph Benchmark (TGB) protocol.} As shown in Table~\ref{tab:tgbl}, several SGNNs outperform the discrete-time baseline ROLAND on tgbl-wiki. However, on the larger tgbl-review dataset, SGNNs achieve inferior performance under limited computational resources. Moreover, both ROLAND and existing SGNNs exhibit substantial performance gaps compared with continuous-time GNNs on both datasets.

These results reveal that most existing SGNNs are designed for node-level tasks, making them less effective on temporal link prediction datasets without node features. More importantly, rate coding-based SGNNs typically process dynamic graphs as discrete snapshots, losing fine-grained temporal information within each snapshot. Although finer temporal bucketing can preserve more temporal information, it substantially increases computational and memory overhead due to the longer snapshot sequences and intermediate spiking states. This presents a dilemma for existing SGNNs on large-scale graphs: balancing fine-grained temporal modeling with computational efficiency. These challenges call for revisiting the event-driven nature of spiking neurons to develop more effective SGNN architectures for relational representation learning.

\begin{table}[htbp]
    \centering
    \caption{Performance comparison on tgbl-wiki and tgbl-review datasets.}
    \label{tab:tgbl}
    \begin{tabular}{lcc}
        \toprule
        \textbf{Models} & \textbf{tgbl-wiki} & \textbf{tgbl-review} \\
        \midrule
        ROLAND    & 0.2803$_{\pm0.0055}$ & 0.2944$_{\pm0.0021}$ \\
        TGAT      & 0.7664$_{\pm0.0029}$ & 0.3356$_{\pm0.0191}$ \\
        DyGFormer & 0.7810$_{\pm0.0013}$ & 0.4106$_{\pm0.0043}$ \\
        TPNet     & 0.8181$_{\pm0.0050}$ & 0.4165$_{\pm0.0025}$ \\
        \rowcolor{gray!20}
        Average   & 0.6615 & 0.3643 \\
        \midrule
        GC-SNN     & 0.2541$_{\pm0.0066}$ & \second{0.2283$_{\pm0.0052}$} \\
        SpikingGCN & \third{0.2879$_{\pm0.0138}$} & >3days \\
        SpikeGT    & \second{0.3240$_{\pm0.0091}$} & \third{0.2119$_{\pm0.0070}$} \\
        SpikeNet   & 0.1631$_{\pm0.0060}$ & 0.2035$_{\pm0.0026}$ \\
        SiGNN      & 0.1373$_{\pm0.0089}$ & >3days \\
        HASNN      & \first{0.3309$_{\pm0.0054}$} & \first{0.2341$_{\pm0.0015}$} \\
        \rowcolor{gray!20}
        Average    & 0.2496 & 0.2195 \\
        \bottomrule
    \end{tabular}
\end{table}

\begin{figure}[bp]
    \centering
    \includegraphics[width=0.9\textwidth]{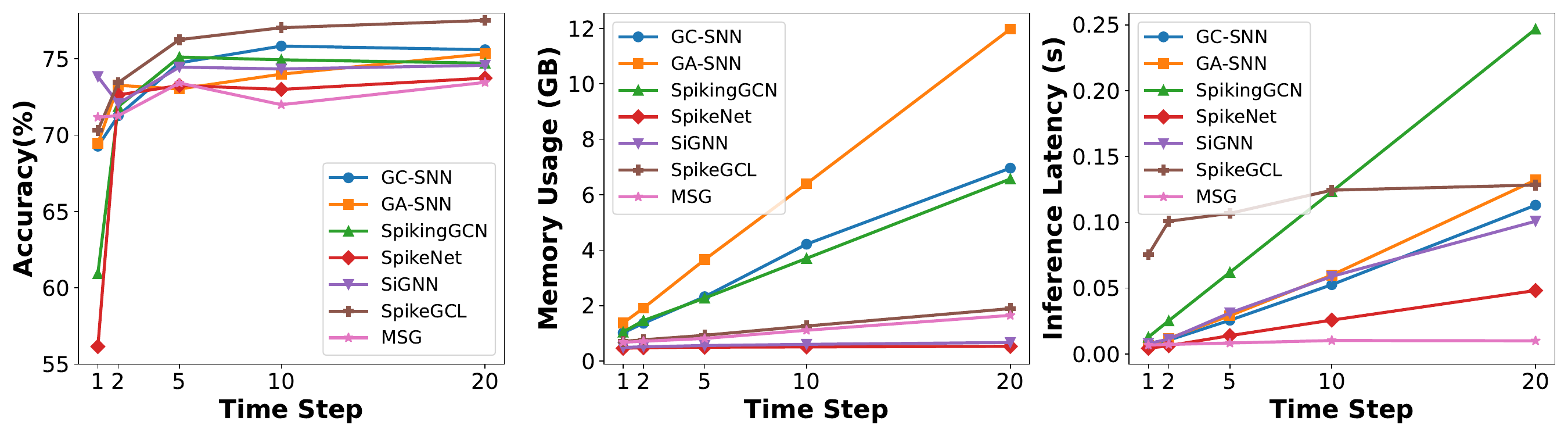}
  \caption{Influence of time step on different SGNN baselines.}\label{fig:2}
\end{figure}

\begin{figure}[bp]
    \centering
    \begin{subfigure}[b]{0.9\textwidth}
        \centering
        \includegraphics[width=\textwidth]{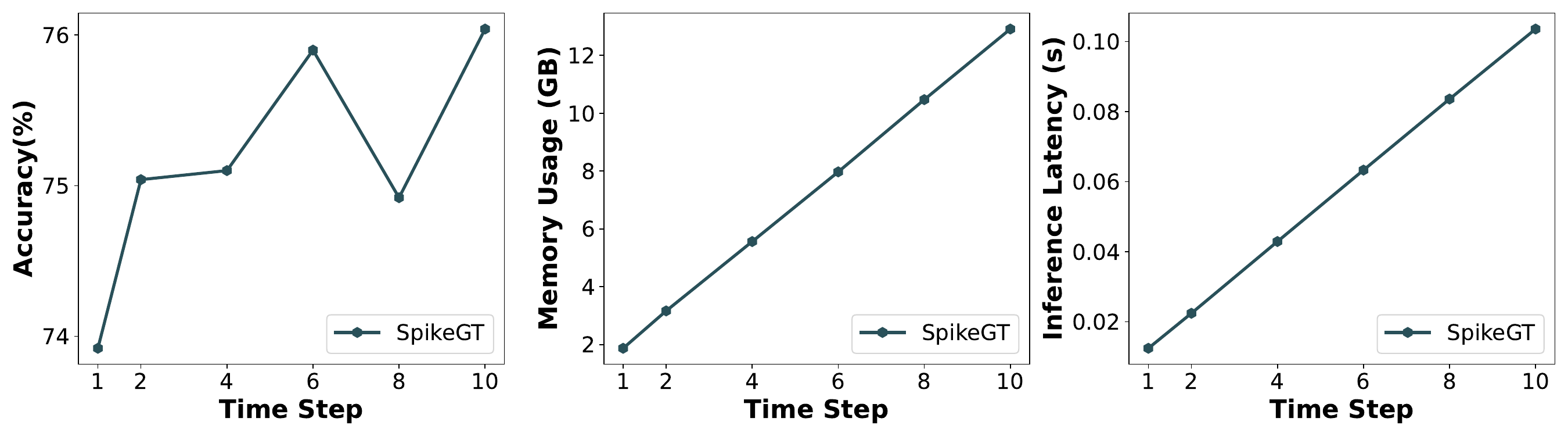}
        \caption{}
    \end{subfigure}
    % \hspace*{0.05\textwidth}
    \begin{subfigure}[b]{0.9\textwidth}
        \centering
        \includegraphics[width=\textwidth]{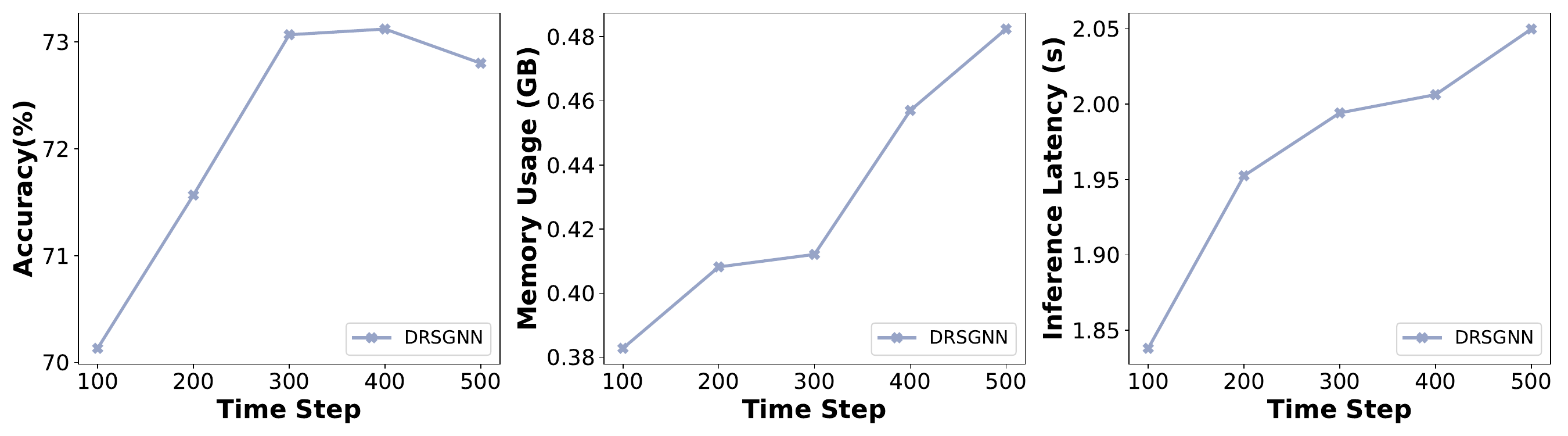}
        \caption{}
    \end{subfigure}
    \caption{Influence of time step on the performance, memory usage, and inference latency of DRSGNN and SpikeGT on PubMed. (a) The time steps of SpikeGT range from 1 to 10. Conversely, (b) DRSGNN requires significantly more time steps (100 to 500) due to its design.}
    \label{fig:3}
\end{figure}

\subsection{Effect of Time Step}
In this section, we deeply investigate the influence of the time step for different SGNNs. Specifically, we fix the other optimal hyperparameter configurations of SGNNs and vary only the time step, while adopting grid search for the learning rate. We repeat each experiment five times on PubMed, and report the average accuracy, maximum memory usage during training and the inference time as evaluation metrics. The visualization results are demonstrated in Figure~\ref{fig:2}. Notably, DRSGNN and SpikeGT operate under distinct time step ranges from other SGNNs owing to their unique implementations. The corresponding results are illustrated in Figure~\ref{fig:3}.

As shown in Figure~\ref{fig:2} and Figure~\ref{fig:3}, most of the SGNNs achieve an acceptable trade-off between performance and efficiency at $T=10$, and their performance tends to plateau by $T=20$. DRSGNN concatenates the positional encoding with node features and feeds them into an MLP Decoder. Its simple network structure exhibits a strong dependency on the number of time steps.
\textbf{For these rate coding-based methods, increasing the number of time steps leads to a higher spike rate, which contributes to training stability.} Conversely, when $T=1$, the extreme sparsity of spikes severely hinders effective training, especially for models such as SpikingGCN, SpikeNet, and DRSGNN.

SNNs typically require multiple time step iterations, which can lead to increased memory usage and slower inference compared to standard neural networks. Visualization results reveal that, for most existing SGNNs, both memory consumption and inference latency grow approximately linearly with the number of time steps. For models like GC-SNN and GA-SNN that incorporate multi-step components, the memory usage tends to increase more rapidly with more time steps. SpikeNet and SiGNN apply neighbor sampling strategies within SGNNs, enabling them to achieve lower memory footprints. Lastly, SpikeGCL is the only method that explicitly discusses designing temporal input sequences over graphs instead of repeatedly feeding the same graph data T times. This design reduces its sensitivity to the number of time steps, allowing it to maintain acceptable memory and inference costs even at larger T values. 

\section{Theoretical Energy Consumption}\label{sup:eng_consumption}

Theoretical energy estimation is widely adopted in energy-efficient deep learning research \cite{horowitz20141}, as actual energy consumption depends heavily on hardware architectures and implementation details. Recent benchmarking efforts, such as NeuroBench \cite{yik2025neurobench}, further distinguish between algorithm-level and system-level evaluations to enable systematic comparisons independent of specific hardware platforms. Following this practice, SGNNBench uses MAC/AC-based theoretical energy estimates for controlled architectural comparisons, abstracting away hardware-specific implementation details.
Here, we formalize the theoretical energy consumption of different SGNNs in Table~\ref{tab:10}. We comprehensively calculate the FLOPs of various SGNN components, including message passing layers ($FP_{MPNN}$), Transformer layers ($FP_{Transformer}$), linear layers ($FP_{Linear}$), normalization schemes ($FP_{BN}$ and $FP_{STFN}$), and positional encoding mechanisms ($FP_{PE}$).

\begin{table}[th]
    \centering
    \caption{Theoretical energy consumption of different SGNNs.}\label{tab:10}
    \begin{tabular}{l|c}
    \toprule
    \textbf{Models} & \textbf{Theoretical energy consumption} \\
    \midrule
    GC-SNN     & \multirow{2}{*}{$E_{ac} \times (FP_{MPNN}+FP_{STFN}) \times T \times R $} \\
    GA-SNN     & \\
    \midrule
    SpikingGCN & \multirow{3}{*}{$E_{ac} \times (FP_{MPNN}+FP_{BN}+FP_{Linear}) \times T \times R $} \\
    SpikeNet   & \\
    MSG        & \\
    \midrule
    SiGNN      &  $E_{ac} \times (FP_{MPNN} \times T + FP_{MPNN} \times \left\lceil T/2 \right\rceil + FP_{MPNN} \times \left\lceil T/3 \right\rceil) \times R $ \\
    \midrule
    SpikeGCL   &  $E_{ac} \times (FP_{GCN\_Blockwise}+FP_{BN\_Blockwise}) \times T \times R $ \\
    \midrule
    SpikeGT    &  $E_{ac} \times FP_{Transformer} \times T \times R + E_{mac} \times FP_{MPNN} $ \\
    \midrule
    DRSGNN     &  $E_{ac} \times (FP_{PE}+FP_{MPNN}+FP_{Linear}) \times T \times R $ \\
    \midrule
    HASNN     &  $E_{ac} \times (FP_{SRGA}+FP_{TA}) \times (T + \left\lceil T/2 \right\rceil) \times R $ \\
    \bottomrule
    \end{tabular}
\end{table}

\subsection{Energy Efficiency Analysis on Dynamic Graphs}
In the aforementioned section, we have compared the energy efficiency of SGNNs and full-precision GNNs on static graphs. Here, we extend this analysis to dynamic graphs, evaluating model size, peak training memory usage and theoretical energy consumption on Patent and tgbl-review. As shown in Table~\ref{tab:energy_dyn}, SGNNs exhibit substantial advantages in theoretical energy efficiency over full-precision dynamic GNNs. On Patent, SpikeNet achieves the lowest theoretical energy consumption of 382.53 mJ, while SpikingGCN achieves the smallest model size and memory footprint. Similarly, on tgbl-review, SpikeNet achieves both the lowest theoretical energy consumption and peak training memory usage. These results highlight the potential of SGNNs for energy-efficient and resource-constrained deployment on large-scale dynamic graphs.

\begin{table}[h]
    \centering
    \caption{The model size (MB), maximum memory usage (GB) and theoretical energy consumption (mJ) on Patent and tgbl-review datasets.}
    \label{tab:energy_dyn}
    \begin{tabular}{l|ccc|ccc}
        \toprule
        \multirow{2}{*}{\textbf{Models}} & \multicolumn{3}{c|}{\textbf{Patent}} & \multicolumn{3}{c}{\textbf{tgbl-review}} \\
        \cmidrule(lr){2-4} \cmidrule(lr){5-7} & \textbf{Size$\downarrow$} & \textbf{Mem$\downarrow$} & \textbf{Eng$\downarrow$} & \textbf{Size$\downarrow$} & \textbf{Mem$\downarrow$} & \textbf{Eng$\downarrow$} \\
        \midrule
        ROLAND    & -    & -     & -         & 0.88   & 12.29 & 4610.90 \\
        TGAT      & 1.51 & 21.11 & 76061.34  & 646.45 & 1.66  & 26552.04 \\
        DyGFormer & 0.88 & 22.47 & 532915.90 & 648.01 & 2.83  & 186033.87 \\
        \midrule
        GC-SNN     & \third{0.17} & \third{1.06} & \second{384.17} & \second{1.41} & 9.23 & \second{41.87} \\
        SpikingGCN & \first{0.10} & \first{0.41} & \third{389.25} & - & - & - \\
        SpikeNet   & \second{0.16} & \second{0.74} & \first{382.53} & \second{1.41} & \first{1.20} & \first{41.69} \\
        SpikeGT    & 1.00 & 2.01 & 3229.11 & \third{2.25} & \third{2.19} & 351.90 \\
        SiGNN      & 0.78 & 1.41 & 2500.48 & - & - & - \\
        HASNN      & 0.90 & 1.13 & 2166.99 & \first{1.18} & \second{1.33} & \third{237.23} \\
        \bottomrule
    \end{tabular}
\end{table}

\section{Additional Ablation Results}\label{sup:extra_ablation}
\begin{wrapfigure}{r}{0.38\textwidth}
  \begin{center}
    \includegraphics[width=\linewidth]{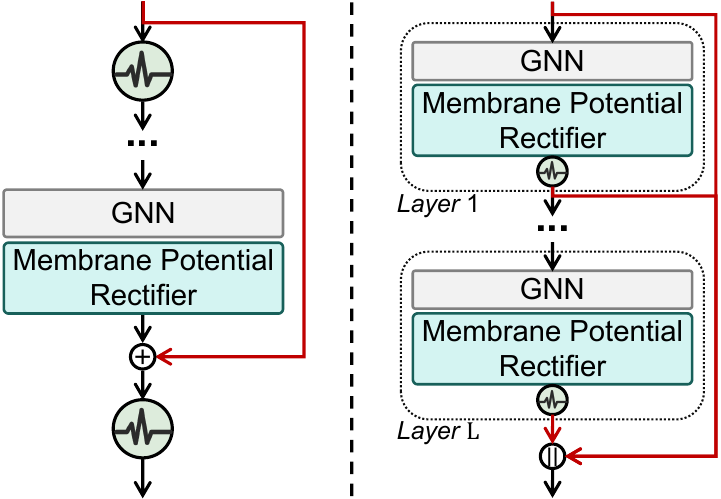}
  \end{center}
  \caption{Membrane Skip Connections (left) and Jumping Knowledge (right) in the ablation study.}\label{fig:4}
\end{wrapfigure}
In Table~\ref{tab:12} and Table~\ref{tab:13}, we provide more ablation results for GC-SNN, GA-SNN and SpikingGCN on 7 homophilic graphs and 6 heterophilic graphs. Besides, Figure~\ref{fig:4} demonstrates that membrane skip connections create shortcuts between the membrane potential of neurons and Jumping Knowledge transfers all of the intermediate spiking representations to the last layer. The ablation results further support and strengthen our claim that \textbf{classical neural network components derived from ANNs, such as MS and JK, effectively improve the performance of SGNNs on both homophilic and heterophilic graphs}. 

On homophilic graphs, adding MS improves the predictive performance of all three architectures on all seven datasets. JK also improves SpikingGCN on all seven datasets, whereas its benefits for GC-SNN and GA-SNN are less uniform. In contrast, PLIF and STFNorm designed for SNNs fail to achieve consistent performance gains across different network architectures on homophilic graphs.
On heterophilic graphs, the effects of the different components vary more substantially across architectures and datasets. Both MS and JK improve GA-SNN on all six datasets. Besides, STFN outperforms BN in only 5 of the 18 settings, highlighting the need for more generalizable membrane potential rectifiers to accommodate complex graphs. PLIF improves performance in 11 of the 18 settings on heterophilic graphs, indicating that spiking neurons with more expressive spiking patterns can produce better spiking representations in most cases.

\begin{table}[t]
    \centering
    \small
    \tabcolsep=1px
    \caption{Ablation studies on homophilic graphs.}\label{tab:12}
    \begin{tabular}{l|c|ccccccc}
        \toprule
        \textbf{Models} & \textbf{Ablation} & \textbf{Cora} & \textbf{CiteSeer} & \textbf{PubMed} & \textbf{Computers} & \textbf{Photo} & \textbf{CS} & \textbf{Physics} \\
        \midrule
        \multirow{5}{*}{GC-SNN} & Vanilla & 80.00$_{\pm0.78}$ & 67.20$_{\pm0.68}$ & 77.35$_{\pm0.65}$ & 91.03$_{\pm0.88}$ & 92.79$_{\pm0.68}$ & 93.68$_{\pm1.71}$ & 96.67$_{\pm0.39}$ \\
        & STFN $\rightarrow$ BN  & 75.00$_{\pm2.40}$ & 62.60$_{\pm2.42}$ & 74.20$_{\pm1.85}$ & 90.59$_{\pm0.44}$ & 88.71$_{\pm2.50}$ & 91.12$_{\pm0.37}$ & 94.58$_{\pm0.21}$ \\
        & LIF $\rightarrow$ PLIF & 82.30$_{\pm1.50}$ & 67.50$_{\pm0.92}$ & 78.22$_{\pm0.65}$ & 91.65$_{\pm0.75}$ & 93.31$_{\pm0.86}$ & 93.65$_{\pm1.55}$ & 96.95$_{\pm0.55}$ \\
        & add MS      & \first{82.96$_{\pm1.03}$} & 68.73$_{\pm1.52}$ & 77.40$_{\pm0.70}$ & \first{93.06$_{\pm3.00}$} & \first{97.02$_{\pm1.51}$} & \first{97.10$_{\pm1.58}$} & \first{97.95$_{\pm0.83}$} \\
        & add JK      & 82.40$_{\pm0.16}$ & \first{69.57$_{\pm0.38}$} & \first{78.83$_{\pm0.70}$} & 91.54$_{\pm0.40}$ & 93.38$_{\pm0.31}$ & 94.08$_{\pm1.47}$ & 96.54$_{\pm0.24}$ \\
        \midrule
        \multirow{5}{*}{GA-SNN} & Vanilla & 76.87$_{\pm4.00}$ & 64.80$_{\pm1.49}$ & 76.10$_{\pm2.16}$ & 88.15$_{\pm0.71}$ & 92.61$_{\pm1.72}$ & 90.31$_{\pm0.41}$ & 95.51$_{\pm0.20}$ \\
        & STFN $\rightarrow$ BN & 64.00$_{\pm2.00}$ & 53.60$_{\pm0.00}$ & 70.00$_{\pm4.09}$ & 88.82$_{\pm1.15}$ & 91.24$_{\pm1.68}$ & 89.66$_{\pm1.02}$ & 94.88$_{\pm0.07}$ \\
        & LIF $\rightarrow$ PLIF & 77.22$_{\pm2.90}$ & 63.44$_{\pm3.44}$ & 74.22$_{\pm2.62}$ & 88.06$_{\pm1.48}$ & 93.69$_{\pm1.00}$ & 90.41$_{\pm0.28}$ & 95.52$_{\pm0.26}$ \\
        & add MS & \first{81.58$_{\pm1.58}$} & \first{68.38$_{\pm2.00}$} & \first{77.42$_{\pm1.99}$} & \first{90.47$_{\pm1.13}$} & \first{96.80$_{\pm1.15}$} & \first{95.11$_{\pm0.55}$} & \first{97.00$_{\pm0.08}$} \\
        & add JK & 80.72$_{\pm2.23}$ & 64.12$_{\pm4.17}$ & 75.04$_{\pm2.32}$ & 88.35$_{\pm0.92}$ & 93.67$_{\pm0.95}$ & 92.17$_{\pm1.95}$ & 95.40$_{\pm0.22}$ \\
        \midrule
        \multirow{5}{*}{SpikingGCN} & Vanilla & 74.40$_{\pm1.35}$ & 63.20$_{\pm1.93}$ & 74.03$_{\pm1.18}$ & 85.64$_{\pm1.01}$ & $92.18_{\pm0.75}$ & 91.71$_{\pm0.24}$ & 95.30$_{\pm0.16}$ \\
        & BN $\rightarrow$ STFN  & \first{80.84$_{\pm1.20}$} & \first{70.70$_{\pm0.67}$} & \first{77.14$_{\pm2.29}$} & 88.06$_{\pm0.68}$ & 92.14$_{\pm0.39}$ & 92.89$_{\pm0.14}$ & 95.20$_{\pm0.42}$ \\
        & LIF $\rightarrow$ PLIF & 78.40$_{\pm1.64}$ & 67.60$_{\pm1.05}$ & 76.40$_{\pm1.85}$ & 91.30$_{\pm0.15}$ & 93.75$_{\pm0.40}$ & 93.36$_{\pm0.48}$ & 96.39$_{\pm0.13}$ \\
        & add MS & 78.52$_{\pm1.69}$ & 64.78$_{\pm1.71}$ & 76.90$_{\pm1.28}$ & \first{91.75$_{\pm0.52}$} & 94.97$_{\pm0.46}$ & 95.25$_{\pm0.29}$ & \first{96.83$_{\pm0.09}$} \\
        & add JK & 78.56$_{\pm2.06}$ & 65.16$_{\pm1.68}$ & 75.44$_{\pm2.42}$ & 91.13$_{\pm0.40}$ & \first{95.39$_{\pm0.52}$} & \first{95.78$_{\pm0.17}$} & 96.15$_{\pm0.12}$ \\
        \bottomrule
    \end{tabular}
\end{table}

\begin{table}[t]
    \centering
    \small
    \tabcolsep=3.0px
    \caption{Ablation studies on heterophilic graphs.}\label{tab:13}
    \begin{tabular}{l|c|cccccc}
        \toprule
        \textbf{Models} & \textbf{Ablation} &\textbf{Squirrel} & \textbf{Chameleon} & \textbf{\makecell{Amazon- \\ Ratings}} & \textbf{\makecell{Roman- \\ Empire}} & \textbf{Minesweeper} & \textbf{Questions} \\
        \midrule
        \multirow{5}{*}{GC-SNN} & Vanilla & \first{43.49$_{\pm1.30}$} & 41.68$_{\pm3.67}$ & 41.82$_{\pm0.28}$ & 55.79$_{\pm0.57}$ & 82.27$_{\pm0.70}$ & 71.36$_{\pm1.10}$ \\
        & STFN $\rightarrow$ BN  & 39.54$_{\pm1.32}$ & 42.77$_{\pm2.70}$ & \first{47.98$_{\pm0.54}$} & 52.95$_{\pm0.38}$ & 82.11$_{\pm0.73}$ & \first{73.81$_{\pm0.39}$} \\
        & LIF $\rightarrow$ PLIF & 43.33$_{\pm1.91}$ & \first{43.37$_{\pm4.52}$} & 42.22$_{\pm0.53}$ & 55.14$_{\pm1.18}$ & 82.38$_{\pm0.80}$ & 73.72$_{\pm0.79}$ \\
        & add MS      & 43.19$_{\pm0.70}$ & 42.94$_{\pm0.69}$ & 44.43$_{\pm0.65}$ & 57.32$_{\pm0.06}$ & 81.80$_{\pm0.66}$ & 73.60$_{\pm0.81}$ \\
        & add JK      & 43.00$_{\pm1.25}$ & 41.73$_{\pm1.95}$ & 42.73$_{\pm0.23}$ & \first{63.62$_{\pm0.52}$} & \first{85.46$_{\pm0.59}$} & \first{73.81$_{\pm0.23}$} \\
        \midrule
        \multirow{5}{*}{GA-SNN} & Vanilla & 40.45$_{\pm0.82}$ & 40.42$_{\pm2.13}$ & 43.83$_{\pm0.64}$ & 65.91$_{\pm1.02}$ & 87.92$_{\pm0.27}$ & 65.51$_{\pm0.39}$ \\
        & STFN $\rightarrow$ BN  & \first{42.45$_{\pm1.56}$} & \first{44.05$_{\pm0.58}$} & \first{48.40$_{\pm0.35}$} & 68.84$_{\pm0.11}$ & 88.10$_{\pm0.42}$ & 72.98$_{\pm0.86}$ \\
        & LIF $\rightarrow$ PLIF & 39.62$_{\pm0.56}$ & 41.07$_{\pm0.53}$ & 42.66$_{\pm0.84}$ & 68.92$_{\pm0.50}$ & 87.75$_{\pm0.81}$ & 69.96$_{\pm0.67}$ \\
        & add MS      & 41.78$_{\pm2.03}$ & 40.50$_{\pm2.22}$ & 45.81$_{\pm1.52}$ & \first{74.06$_{\pm0.82}$} & \first{90.36$_{\pm0.37}$} & \first{74.27$_{\pm1.12}$} \\
        & add JK      & 40.75$_{\pm0.98}$ & 41.42$_{\pm3.13}$ & 46.02$_{\pm0.32}$ & 68.00$_{\pm0.70}$ & 88.68$_{\pm0.40}$ & 69.03$_{\pm1.22}$ \\
        \midrule
        \multirow{5}{*}{SpikingGCN} & Vanilla & 41.49$_{\pm0.87}$ & 40.09$_{\pm1.65}$ & \first{49.47$_{\pm0.49}$} & 65.23$_{\pm1.04}$ & 74.17$_{\pm1.13}$ & \first{76.18$_{\pm0.58}$} \\
        & BN $\rightarrow$ STFN  & 41.62$_{\pm1.24}$ & 42.49$_{\pm2.24}$ & 46.81$_{\pm0.22}$ & 61.74$_{\pm0.20}$ & 74.08$_{\pm0.96}$ & 72.23$_{\pm0.56}$ \\
        & LIF $\rightarrow$ PLIF & 43.27$_{\pm0.28}$ & 42.50$_{\pm2.95}$ & 47.56$_{\pm0.47}$ & 67.72$_{\pm1.05}$ & 74.32$_{\pm0.95}$ & 74.08$_{\pm1.47}$ \\
        & add MS      & \first{43.79$_{\pm0.18}$} & \first{46.68$_{\pm1.70}$} & 48.16$_{\pm0.62}$ & \first{72.09$_{\pm0.62}$} & \first{90.76$_{\pm0.72}$} & 75.65$_{\pm0.85}$ \\
        & add JK      & 43.12$_{\pm0.96}$ & 42.76$_{\pm2.73}$ & 46.89$_{\pm0.54}$ & 65.92$_{\pm0.80}$ & 75.96$_{\pm1.16}$ & 72.38$_{\pm1.05}$ \\
        \bottomrule
    \end{tabular}
\end{table}

\subsection{Significance Test}\label{sup:significance_test}

\begin{wrapfigure}{r}{0.45\textwidth}
  \vspace{-10px}
  \begin{center}
    \includegraphics[width=\linewidth]{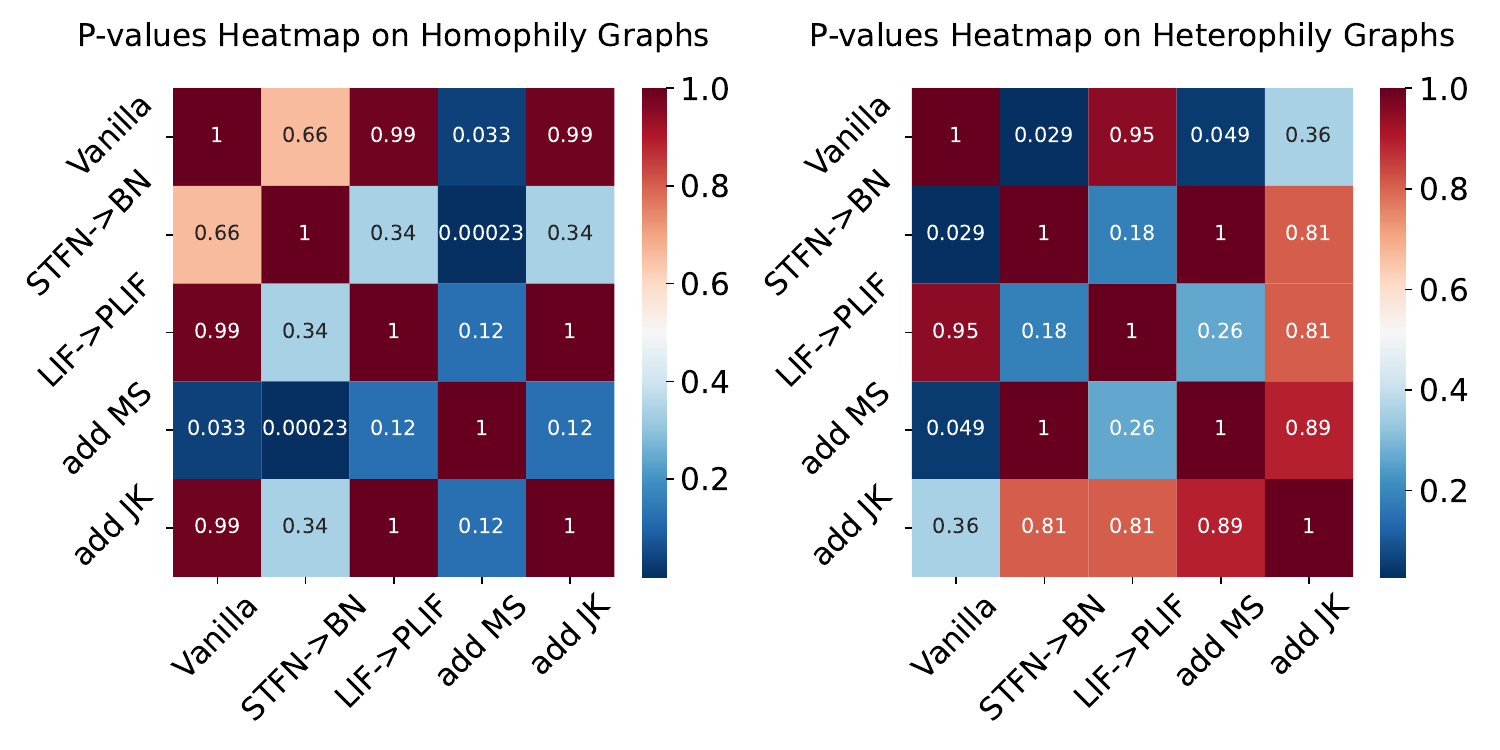}
  \end{center}
  \caption{The significance tests on homophilic and heterophilic graphs.}\label{fig:5}
  \vspace{-10px}
\end{wrapfigure}

We employ the Nemenyi test \cite{nemenyi1963distribution} to evaluate the performance differences among GA-SNN variants with different components across multiple datasets. The heatmaps of p-values on 7 homophilic and 6 heterophilic graphs are shown in Figure~\ref{fig:5}. 
The results indicate that \textbf{models with the MS significantly outperform the original model on both homophilic and heterophilic graphs}. Additionally, \textbf{on heterophilic graphs, models using the Batch Normalization technique show statistically significant improvements over the original version}.

Here, we provide our analysis that this inconsistency stems from the computations of the mean and standard deviation. Traditional GNNs apply batch normalization to 2D node representations, essentially computing channel-wise statistics across all nodes. In contrast, STFNorm computes the mean and standard deviation along the feature and temporal dimensions. Given the number of features $C$, the number of nodes $N$, and the number of time steps $T$, the mean and variance in STFNorm can be formulated as $E[x]=\frac{1}{CT}\sum_{t}^{T}\sum_{c}^{C}x_c^t, \ Var[x]=\frac{1}{CT}\sum_{t}^{T}\sum_{c}^{C}(x_c^t - E[x])^2$.
This operation is more akin to LayerNorm, which can partially mitigate the undesired distribution shift of membrane potentials. However, prior work has shown that LayerNorm brings little improvement to the training process in certain tasks, which may explain the inconsistent effectiveness of STFNorm across different SGNN architectures \cite{pmlr-v139-cai21e, you2021designspacegraphneural}.

\section{Limitation}\label{sup:limitation}
Although this work provides a unified framework, implementing and training spiking neural networks remains fundamentally more complex than artificial neural networks. In graph tasks, progress in key techniques such as surrogate gradient design, time step scheduling, and threshold tuning is still limited. Moreover, most existing graph-based spiking neural networks rely on the rate coding scheme. A few methods based on temporal coding are primarily restricted to knowledge graph completion or link prediction tasks, making it difficult to include them in SGNNBench for holistic comparison. 

\section{Related Work}\label{sup:related_work}
\paragraph{Energy-Saving GNNs.}
GNNs have demonstrated significant success in handling graph-structured data due to their ability to model complex relationships and interactions. However, the existing inference and training strategies in classic GNNs pose challenges for deploying GNNs on large-scale and complex graphs in resource-constrained environments, such as mobile devices or edge computing platforms.
For developing energy-saving GNNs, some studies focus on graph sampling strategies \cite{xu2018powerful, hamilton2017inductive}. For example, PinSAGE \cite{ying2018graph} introduces the random walk algorithm to generate importance scores for neighbors to select the key neighborhood information.
Besides, GNN Quantization, which maps continuous data like neural network weights or activations to more compact representations, significantly reduces the inference energy consumption of GNNs. 
Typically, this process involves converting high-precision numerical values to lower precision formats. For example, Bi-GCN \cite{wang2021bi} proposes a binary GCN, which replaces heavy matrix multiplications with bitwise operations like XNOR or Bitcount. 

\paragraph{Spiking Neural Networks.} Spiking Neural Networks (SNNs), inspired by brain-like spiking computation, are designed to reduce energy consumption in neural network models. Unlike Artificial Neural Networks (ANNs), SNNs use binary and sparse spikes for communication, which reduces the storage cost of intermediate embeddings and relies more on addition operations rather than multiplication \cite{roy2019towards}.
Motivated by these energy-saving benefits, researchers have explored two main approaches to building spike-driven networks: ANN-to-SNN conversion and direct training.
ANN-to-SNN conversion methods build SNNs from pre-trained ANNs. They reduce information loss during conversion by scaling weights or replacing activation functions with spiking neurons \cite{diehl2015fast, cao2015spiking, hao2023reducing}. For the latter, many studies try to directly train spike-driven neural networks via the surrogate gradient, which alleviates the strong reliance on time steps \cite{fang2021deep, zheng2021going}.

\paragraph{Spiking Graph Neural Networks.} Spiking Graph Neural Networks (SGNNs) combine Spiking Neural Networks (SNNs) with Graph Neural Networks (GNNs) to enable energy-efficient graph learning by mimicking the brain's event-driven computation. Instead of continuous-valued activations, SGNNs use discrete spike signals to propagate information through the graph, reducing redundant computations and power consumption. The event-driven message passing means nodes only update when they receive significant signals, making them well-suited for neuromorphic hardware and real-time processing of dynamic graphs. A line of research shows that SGNNs enhance energy efficiency while maintaining the expressive power of traditional GNNs \cite{dold2021spike, xu2023exploiting, snyder2024transductive}.

\end{document}